\documentclass[pdflatex,sn-basic,iicol]{sn-jnl}% Basic Springer Nature Reference Style/Chemistry Reference Style
\usepackage{graphicx}%
\usepackage{multirow}%
\usepackage{amsmath,amssymb,amsfonts}%
\usepackage{amsthm}%
\usepackage{mathrsfs}%
\usepackage[title]{appendix}%
\usepackage{xcolor}%
\usepackage{textcomp}%
\usepackage{manyfoot}%
\usepackage{booktabs}%
\usepackage{algorithm}%
\usepackage{algorithmicx}%
\usepackage{algpseudocode}%
\usepackage{listings}%
\usepackage{caption}    % optional for caption tweaking
\usepackage{microtype}      % microtypography
\usepackage{setspace}
\usepackage{multicol}
\usepackage{makecell}
\usepackage{color}
\usepackage{tikz}
\usepackage{enumitem}
\usepackage{subcaption}
\usepackage{diagbox}
\usepackage{float}      % 更精确的浮动控制

\theoremstyle{thmstyleone}%
\theoremstyle{thmstyletwo}%

\theoremstyle{thmstylethree}%

\begin{document}

\title[Article Title]{GLAD: Generative Language-Assisted Visual Tracking for Low-Semantic Templates}


\author[1]{\fnm{} \sur{Xingyu Luo}}\email{652025330020@smail.nju.edu.cn}
\equalcont{These authors contributed equally to this work.}

\author[1]{\fnm{} \sur{Yidong Cai}}\email{dawnyc1123@gmail.com}
\equalcont{These authors contributed equally to this work.}

\author*[1]{\fnm{} \sur{Jie Liu}}\email{liujie@nju.edu.cn}

\author[1]{\fnm{} \sur{Jie Tang}}\email{ tangjie@nju.edu.cn}

\author[1]{\fnm{} \sur{Gangshan Wu}}\email{gswu@nju.edu.cn}

\author[1]{\fnm{} \sur{Limin Wang}}\email{lmwang@nju.edu.cn}

\affil[1]{\orgdiv{State Key Laboratory for Novel Software Technology}, \orgname{Nanjing University}, \orgaddress{\city{Nanjing}, \postcode{210023}, \country{China}}}


\abstract{Vision-language tracking has gained increasing attention in many scenarios. This task simultaneously deals with visual and linguistic information to localize objects in videos. Despite its growing utility, the development of vision-language tracking methods remains in its early stage.
Current vision-language trackers usually employ Transformer architectures for interactive integration of template, search, and text features. However, persistent challenges about low-semantic images including prevalent image blurriness, low resolution and so on, may compromise model performance through degraded cross-modal understanding.
To solve this problem, language assistance is usually used to deal with the obstacles posed by low-semantic images. However, due to the existing gap between current textual and visual features, direct concatenation and fusion of these features may have limited effectiveness.
To address these challenges, we introduce a pioneering \textbf{G}enerative \textbf{L}anguage-\textbf{A}ssiste\textbf{D} tracking model, \textbf{GLAD}, which utilizes diffusion models for the generative multi-modal fusion of text description and template image to bolster compatibility between language and image and enhance template image semantic information. Our approach demonstrates notable improvements over the existing fusion paradigms. Blurry and semantically ambiguous template images can be restored to improve multi-modal features in the generative fusion paradigm. Experiments show that our method establishes a new state-of-the-art on multiple benchmarks and achieves an impressive inference speed. The code and models will be released at: https://github.com/Confetti-lxy/GLAD}


% \pacs[JEL Classification]{D8, H51}


\maketitle

\section{Introduction}\label{sec: introduction}

Vision-language tracking task~\citep{feng2021siamese, wang2021towards, zhou2023joint, guo2022divert} has gradually enabled various applications in daily life scenarios. This task usually requires descriptive ~\citep{zheng2023toward} information from both the vision and language modality of the target object. Vision-language trackers will generate a bounding box prediction as the output on each video frame. 
Considering that natural language can provide clear semantic information~\citep{yang2019fast, huang2021look} while visual templates can depict the textural information~\citep{cui2022mixformer, ye2022joint}, unification of these two kinds of descriptive information will help reduce the ambiguity during tracking and finally assist the tracker to recognize the target. Unfortunately, research on vision-language tracking remains limited at present, particularly in scenarios involving low-semantic images.

Up to the present moment, for the traditional visual tracking task~\citep{Staple, BhatJDKF18, ATOM, Song0WGBZSL018, WangZLWTB21, DBLP:conf/cvpr/ChenPWLH23, wei2023autoregressive}, the descriptive information is usually a template image which contains the target object. Meanwhile, the tracking by natural language specification"~\citep{li2017tracking, feng2020real} attempts to locate the template only by the text initialization and then construct their pipeline similar to visual tracking. Some vision-language trackers~\citep{guo2022divert, zhou2023joint} borrow ideas from natural language tracking methods to perform separate processing on texts and then perform integration with visual features for enhancement.

\begin{figure}[htbp]
\centering
\includegraphics[width=0.99\columnwidth]{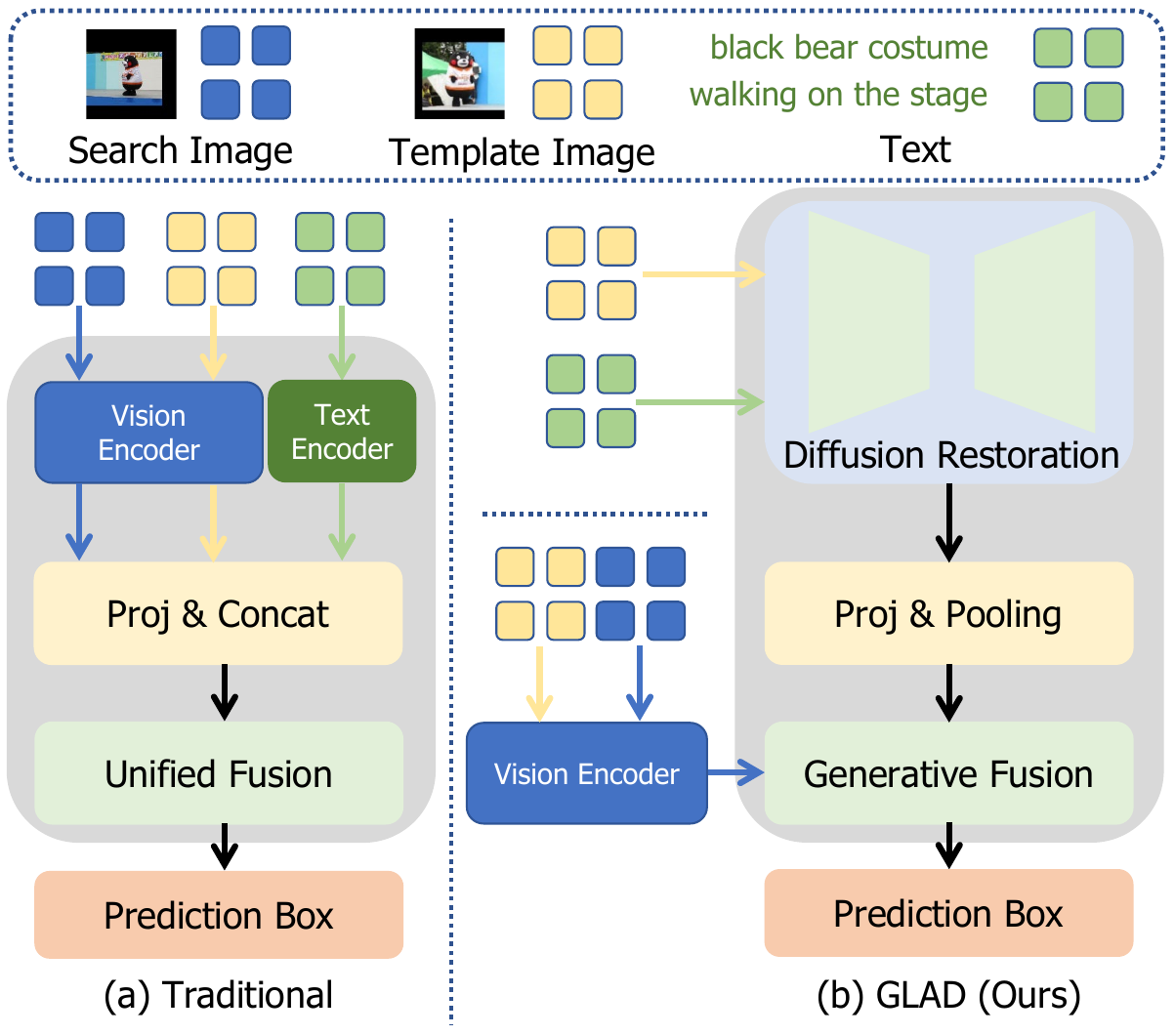}
\caption{Paradigms for Vision-Language trackers. The top section shows the inputs, including text, search image and template image. (a) Traditional paradigm on the left directly concatenations different features and conducts a simple unified fusion. (b) Our generative paradigm on the right employs diffusion models for restoration of suboptimal template images, and interactions with original images are then conducted during generative fusion.
}
\label{Fig:introduction}
\end{figure}

The classical vision-language trackers, as shown on the left half of Figure~\ref{Fig:introduction}, mainly extract features from different modalities based on separate networks. Generally, text features are extracted by text models such as BERT~\citep{devlin2018bert} and GPT~\citep{brown2020language}, while image features, including template and search features, are extracted in a siamese way from image models such as ResNet~\citep{he2016deep} and ViT~\citep{dosovitskiy2020image}. After that, different multi-modal fusion schemas will be taken to enhance image features with text features in different paradigms. 

\begin{figure}[htbp]
\centering
\includegraphics[width=0.99\columnwidth]{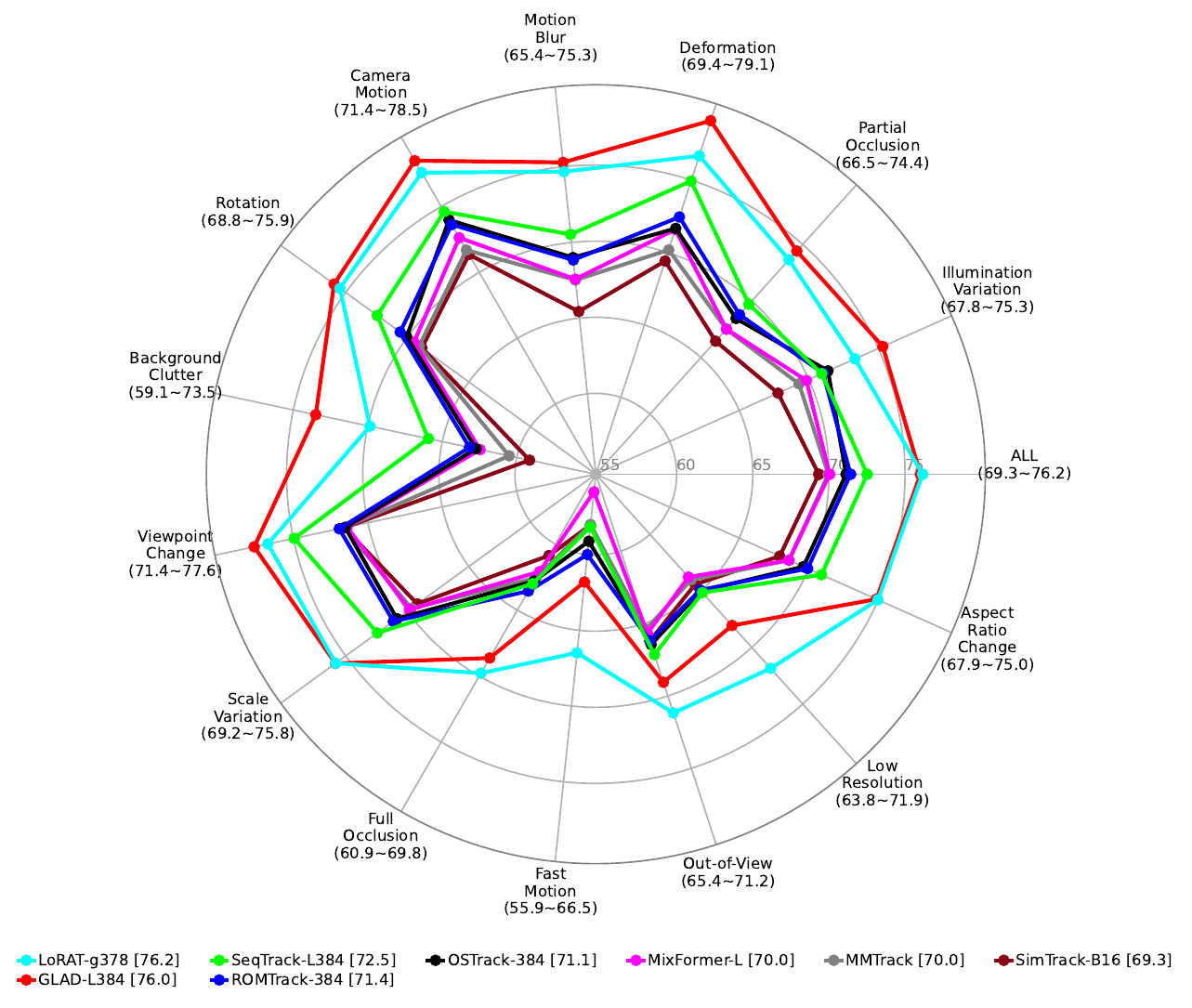}
\caption{Radar plot of AUC scores for LaSOT attributes, including Motion Blur, Camera Motion, Deformation, Illumination Variation, Out-of-View, Scale Variation, and Background Clutter. 
The figure illustrates the tracking performance of different models under diverse challenging conditions. 
A larger enclosed area indicates better overall tracking performance across all attribute types.
}
\label{fig:radar}
\end{figure}

However, these paradigms extract text features and send them directly into fusion stage, connections between text and visual representations are not sufficiently considered for this task. 
In cases of low-semantic templates, which we consider from the low clarity of the tracking target and the weak relevance to the text description, 
feature fusion is further hindered, which adversely affects the prediction results and leads to poor performance.
Besides, the inherent limitations of existing vision-language tracking models, which are predominantly discriminative in nature, relying on feature interaction between the text, template, and search region to generate the final bounding box. As a result, these models typically lack the generative capability to reconstruct or enhance input data. 

In light of these challenges, we propose the generative fusion paradigm shown on the right half of Figure~\ref{Fig:introduction}, which focuses on leveraging a generative fusion to bolster compatibility between language and image. In contrast of discriminative models, generative models like Stable Diffusion possess stronger reconstruction abilities and can produce features during the generation process that are otherwise inaccessible to discriminative approaches. These features can act as a form of guidance, helping the model better understand and locate the target object. Specifically, text and template features are fused in a generative schema, enhancing the semantic fusion process to improve the semantic richness of low-semantic templates (visualized in Section~\ref{sec:vision}). Compared with existing paradigms, our paradigm prevents semantic mismatches between texts and search images, restores the template and substitutes the process by interaction with diffusion features. Diffusion features are more robust because the generative fusion process augments details of the target in template features with the help of texts, thus promoting the relation modeling process.

Based on this paradigm, we propose a new vision-language tracking method GLAD, which employs diffusion models to perform generative multi-modal fusion. Verified by experiments, our method surpasses most existing state-of-the-art vision-language trackers in terms of both tracking accuracy and inference speed on LaSOT~\citep{fan2019lasot}, LaSOT$_{\text{ext}}$~\citep{fan2021lasot},  TNL2K~\citep{wang2021towards} and OTB99-lang~\citep{li2017tracking}.
In particular, our method exhibits significant advantages in the attributes of background clutter and deformation, showcasing a superior discriminative ability of the model, which is shown in Figure~\ref{fig:radar}.

In general, our contributions are three-folds: (1) We propose a new generative paradigm for vision-language tracking which augments template features with the help of text, offering a novel perspective to the field. (2) Building upon this paradigm, we propose a new vision-language tracking framework named GLAD, which can effectively restore low-semantic templates. (3) Our GLAD-L384 model achieves state-of-the-art results on several vision-language tracking datasets and also surpasses many vision-only models, maintaining an impressive inference speed of 44 FPS.

\section{Related Work}\label{sec: related}

\subsection{Vision-Language Tracking}

The core of the vision-language tracking task~\citep{li2017tracking, feng2020real, yang2020grounding, li2022cross, guo2022divert, shao2024context} lies in effectively leveraging both visual and textual cues of a specific target to achieve accurate localization in a given video sequence. Compared to traditional vision-only trackers, which rely solely on appearance information extracted from image sequences, vision-language trackers introduce additional semantic guidance from natural language descriptions. This allows the model to better distinguish between visually similar objects and to remain robust under challenging scenarios such as occlusion, illumination variation, and background clutter. In this setting, the textual description provides a complementary semantic reference for the visual features, thereby serving as a high-level constraint that helps disambiguate the target. As a result, the central challenge of vision-language tracking becomes how to conduct effective and interpretable multi-modal fusion between visual and linguistic representations.

Most existing solutions~\citep{feng2021siamese, wang2021towards} employ pre-trained language encoders such as BERT~\citep{devlin2018bert} and GPT~\citep{brown2020language} to embed textual features into vectorized tokens. These token embeddings are then incorporated into visual features through attention mechanisms or feature modulation strategies, enabling interaction between the two modalities. Transformer-based architectures~\citep{vaswani2017attention} have become the mainstream framework for such multi-modal fusion due to their strong capacity for contextual reasoning. Representative examples include TransVLT~\citep{zhao2023transformer}, which employs a cross-attention mechanism to align visual and linguistic embeddings, and JointNLT~\citep{zhou2023joint}, which jointly models target grounding and tracking through a unified Transformer framework.

Since the current multi-modal fusion design is naive and inflexible, we propose a novel generative multi-modal fusion schema to solve the problem.

\subsection{Diffusion Model}

Diffusion models~\citep{ho2020denoising, sohl2015deep, song2019generative, song2020score, DBLP:conf/cvpr/LugmayrDRYTG22} have emerged as the new state-of-the-art family of generative models in computer vision. The key idea of diffusion models like DDPMs~\citep{ho2020denoising, sohl2015deep}, SGMs~\citep{song2019generative, song2020improved} and Score SDEs~\citep{karras2022elucidating, song2021maximum, song2020score} is to progressively perform denoising process on noised data. For example, DDPM mainly uses two Markov chains: a forward chain that adds noise to images and a reverse chain that converts noised data back to images. 

With the advancement of large-scale vision-language models, diffusion models have been further extended to cross-modality generation tasks, as exemplified by Stable Diffusion~\citep{rombach2022high} and DALL\!·\!E~2~\citep{rassin2022dalle}. These approaches enable the generation of high-fidelity visual content conditioned on textual prompts, greatly promoting the development of Artificial Intelligence Generated Content (AIGC). 
Inspired by their powerful generative capability, many subsequent works have explored diffusion-based architectures for image restoration tasks, focusing on reconstructing fine-grained textures and recovering degraded visual details, which in turn enhance the perceptual realism and semantic consistency of the restored images. ~\citep{li2025diffusion, zhao2020layout2image}
Such improvements in image quality not only lead to visually pleasing results but also provide more reliable and semantically complete representations.~\citep{russakovsky2015imagenet, quan2024deep}

Recently, many other methods, such as efficient sampling and improved likelihood, have been applied to diffusion models to improve inference speed. Furthermore, some studies~\citep{chen2023diffusiondet, pnvr2023ld} have demonstrated that the appropriate integration of generative diffusion models into the denoising process can provide auxiliary structural and semantic cues, thereby benefiting a range of downstream vision tasks~\citep{cheng2025breaking} such as object detection~\citep{ge2025improving} and ultimately enhancing the representational capacity and robustness of visual models.
Building on this insight, we propose a generative multimodal fusion schema that incorporates diffusion-based features into the tracking process, aiming to overcome the rigidity of current discriminative fusion designs.

\begin{figure*}[htbp]
    \centering
    \begin{subfigure}[b]{0.58\textwidth}
        \includegraphics[width=\textwidth]{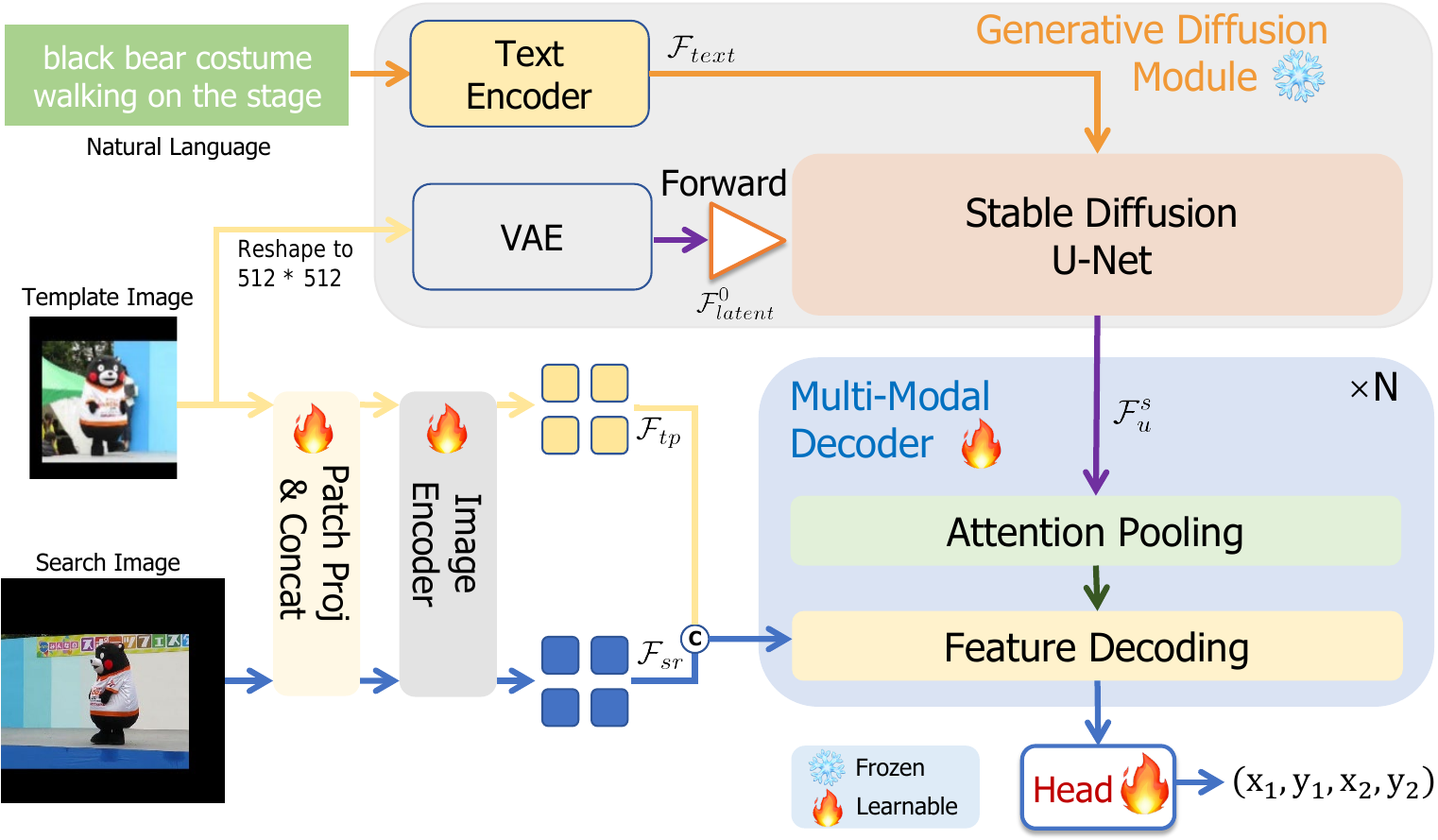} 
        \caption{Pipeline of GLAD}
        \label{fig:pipeline}
    \end{subfigure}
    \hspace{2pt}
    \begin{tikzpicture}[baseline]
        \draw[dashed, thick] (0,0) -- (0,6.8);
    \end{tikzpicture}
    \hspace{2pt}
    \begin{subfigure}[b]{0.38\textwidth}
        \begin{subfigure}[b]{\textwidth}
            \includegraphics[width=\textwidth]{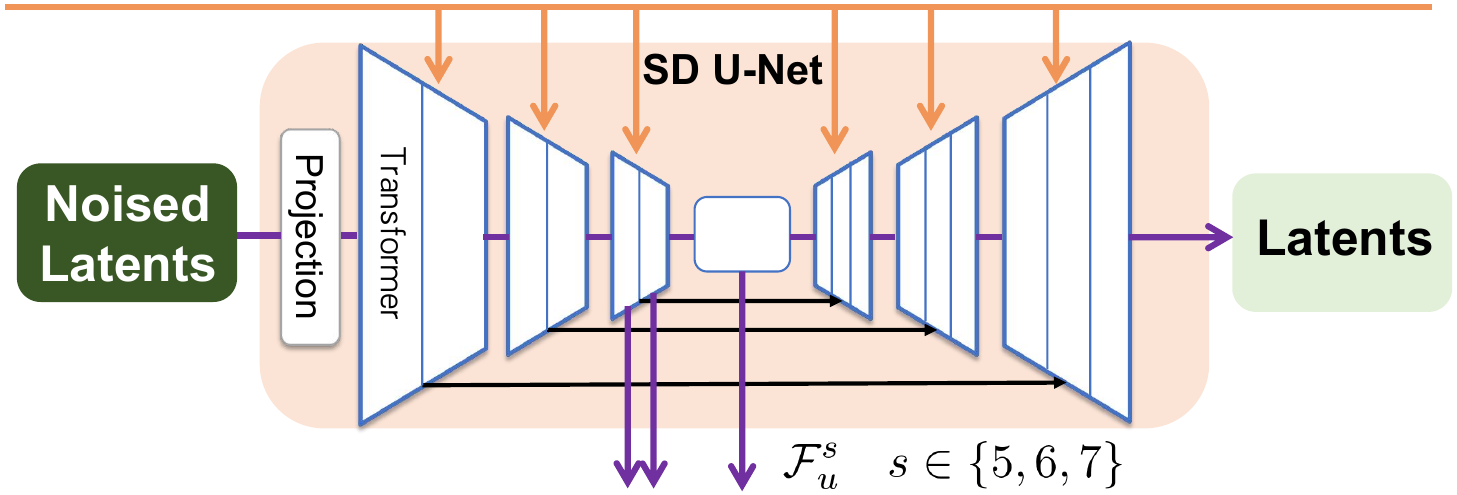} 
            \caption{Stable Diffusion U-Net}
            \label{fig:stable diffusion u-net}
        \end{subfigure}
        \begin{subfigure}[b]{\textwidth}
            \hspace{11pt}
            \includegraphics[width=0.85\textwidth]
            {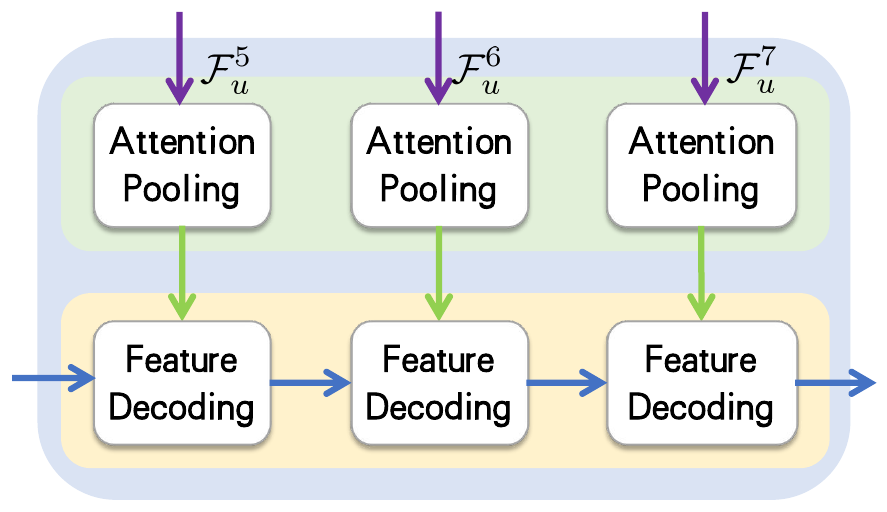} 
            \caption{Multi-Modal Decoder ($N=3$)}
            \label{fig:multi-modal decoder}
        \end{subfigure}
    \end{subfigure}
    \caption{Pipeline of the proposed GLAD model. (a) Firstly, the natural language and the template image will be generatively fused using the Generative Diffusion Module. Secondly, the generated diffusion features are passed through the cascaded Multi-Modal Decoder to enhance semantic information. Search features are concatenated with template features for decoding modules under the guidance of the fused features from pooling modules. (b) Details of Stable Diffusion U-Net. (c) Details of Multi-Modal Decoder when $N=3$.}
    \label{fig:GLAD}
\end{figure*}
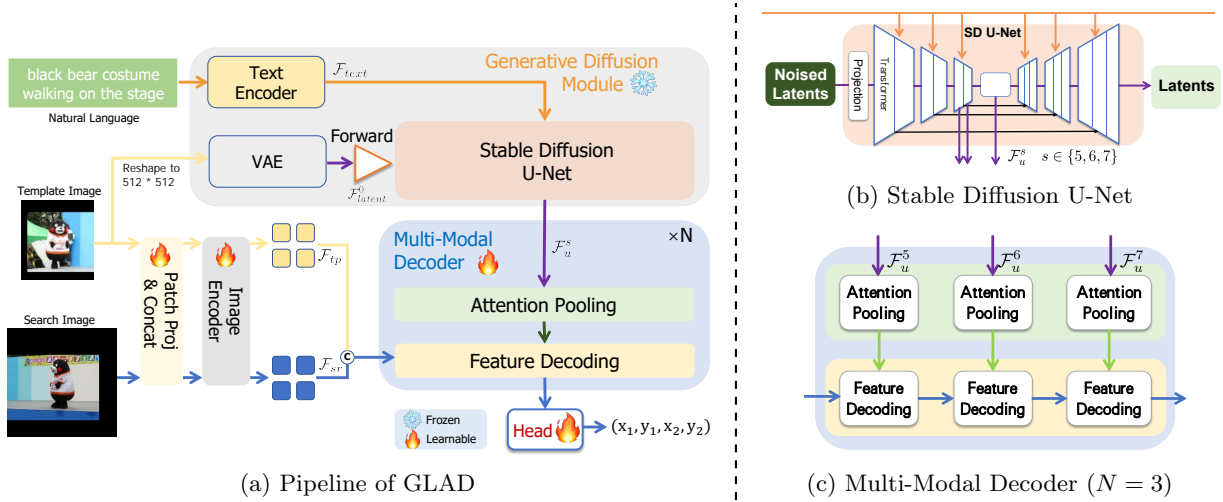

\section{Method}\label{sec: method}

For the tracking task, the model needs to determine the target’s position in each search frame based on both the template image and the accompanying text description. Although the template provides essential visual cues, it often lacks sufficient semantic information, which increases the difficulty of distinguishing the target in cluttered or ambiguous scenes. To compensate for this limitation, textual descriptions are introduced to supply richer semantic context and guide the tracking process. Nevertheless, due to the inherent semantic gap between textual and visual representations, directly incorporating language features into visual pipelines usually provides only limited improvements, especially when the visual template itself is of low quality or poorly aligned with the textual description.

Therefore, we intend to use a diffusion model to leverage text information for template restoration and obtain a highly semantic fused feature, thereby enhancing the performance of vision-language trackers. As a result, we introduce diffusion model in our method that employs text to guide and rectify the template. 

Figure~\ref{fig:pipeline} shows the overall framework of GLAD, including three main components: the Generative Diffusion Module, the cascaded Multi-Modal Decoders ($N$ in total, each consists of one attention pooling module and one feature decoding module), and the Head used to predict the final coordinate. 

\subsection{Generative Diffusion Module}
\label{GDM}

Due to the inherent semantic gap between textual and visual features, traditional fusion paradigms that directly combine text and template representations often fail to make full use of the semantic richness of language, treating textual cues as auxiliary signals rather than deeply integrated guidance. This limitation is particularly evident in discriminative models, which mainly rely on feature interaction without the capacity to reconstruct or refine the target representation. In contrast, generative diffusion models have recently demonstrated remarkable success in both image generation and video understanding, showing their ability to capture fine-grained semantics and structural details during the denoising process. Such generative capabilities offer a promising new approach for enhancing template quality and bridging the semantic gap in vision-language tracking.

By leveraging the image-to-image capability of diffusion models, text descriptions and template images can be semantically combined to create high-quality template images and effectively utilize language information, which we refer to as template inpainting. This novel process will restore the template image and thus improve low-semantic template. Using the restored template from inpainting is unnecessary because we can simplify by leveraging intermediate features, whereby text and template features are fused to yield semantically richer multi-modal features. As shown in Figure~\ref{fig:stable diffusion u-net}, the proposed method employs Stable Diffusion (SD) for generative fusion. It is based on the Latent Diffusion Model (LDM)~\citep{song2019generative}, which employs the similar of DDPM~\citep{ho2020denoising, sohl2015deep} in latent space. 

Specially, DDPM generally uses two Markov chains, the forward one to perturb data to noise and the reverse one to remove this noise. For generating new data samples, DDPM starts by generating an unstructured noise vector from the prior distribution, then gradually removing noise by running a learnable Markov chain in the reverse process. Formally, DDPMs can be interpreted as a sequence of denoising autoencoders \(\epsilon_{\theta}(x_t, t);t=1,\ldots T\), which are trained to predict a denoised variant of their input \(x_t\), where \(x_t\) is a noisy version of the input \(x\). The corresponding objective can be simplified to:
\begin{equation}
    L_{DDPM} = E_{x, \epsilon \sim \mathcal{N}(0, 1), t}\left[\| \epsilon - \epsilon_{\theta}(x_t, t) \|_{2}^{2} \right],
\end{equation}
where \(E\), \(\mathcal{N}\) denotes the normal distribution and \(t\) is uniformly sampled from \(\left\{1, \ldots T \right\}\).

Compared to DDPM, LDM almost does the same thing, the difference is just that LDM needs to transform the high-dimensional pixel space into an efficient, low-dimensional latent space using the VAE~\citep{kingma2013auto, rezende2014stochastic} encoder, where we will apply DDPM. The objective on the perceptually most relevant bits uses the reweighted bound, which now reads:
\begin{equation}
    L_{LDM} = E_{\mathcal{E}(x), \epsilon \sim \mathcal{N}(0, 1), t}\left[\| \epsilon - \epsilon_{\theta}(z_t, t) \|_{2}^{2} \right].
\end{equation}
The neural backbone \(\epsilon_{\theta}(\circ, t)\) of the diffusion model is implemented as a time-conditional U-Net~\citep{dhariwal2021diffusion, ho2020denoising, ronneberger2015u, song2020score}. Latent features of samples can also be decoded to image space using the VAE decoder.

In our GLAD pipeline, the U-Net from SD serves as the core architecture. It accepts noised latents and outputs denoised latents which are semantically restored. Template images are transferred to the latent space using VAE~\citep{kingma2013auto, rezende2014stochastic}, and then altered to noised latents through the forward process:
\begin{equation}
    \left\{
    \begin{array}{l}
    \begin{aligned}
        \mathcal{F}_{latent} &= \text{VAE}(template)\\
        \mathcal{F}^{0}_{latent} &= \text{Forward}(\mathcal{F}_{latent})
    \end{aligned}
    \end{array}
    \right.,
\end{equation}
where $template$ represents the input template image and $\mathcal{F}^{0}_{latent}$ represents the noised latents.

As shown in Figure~\ref{fig:stable diffusion u-net}, the U-Net structurally comprises two parts: the left contracting path and the right expanding path. The contracting path experiences a gradual decrease in resolution and increases in dimension, with the level of text information deepening, leading to progressively richer multi-modal semantic features. The expanding path, on the other hand, corresponds to an opposite upsampling process, with deep-layer features being used to recover shallow-layer information, thus aligning semantics more closely with the original latent representation. 
The U-Net in Figure~\ref{fig:stable diffusion u-net} consists of 16 Transformer submodules, with 7 in the contracting path and 9 in the expanding path:
\begin{equation}
    \left\{
    \begin{array}{l}
    \begin{aligned}
        \mathcal{F}_{text} &= \text{TextEncoder}(text)\\
        \mathcal{F}^s_{u} &= \text{Transformer}_s (\mathcal{F}^{s-1}_{latent}, \mathcal{F}_{text})
    \end{aligned}
    \end{array}
    \right.,
\end{equation}
where $s\in \{1,2,3,...,16\}$ represents the sequence number of Transformer submodules and $text$ represents the input natural language. Besides, $\mathcal{F}^{16}_{u}$ represents the output denoised latents.

Aiming to fully exploit the deep semantic representations while still benefiting from shallower fusion cues and maintaining inference efficiency, we select three submodules from the contracting path of the U-Net, namely the 5-th, 6-th, and 7-th layers. The corresponding feature maps, denoted as $\mathcal{F}^{s}_{u} \quad (s \in {5,6,7})$, are then transferred to the Multi-Modal Decoder for further processing. These layers are chosen because they capture increasingly abstract semantic information while retaining sufficient spatial structure, thus striking a balance between expressiveness and computational overhead. By incorporating the U-Net features in this manner, textual information can be more effectively integrated into the template, enabling the restoration of higher-quality target representations and alleviating the limitations of direct feature fusion.

Notably, this design does not amplify the discrepancy nor degrade performance, even when reshaping templates are used. This is because the template frame does not directly interact with the search image; instead, it is processed solely by the diffusion model, and only the resulting multimodal, target-centric semantic features are involved in the interaction, which prevents mismatches from affecting tracking performance.

As for the design choice of not directly feeding the generated template image into the backbone, while using the generated template image as input to the backbone could potentially enable stronger interaction with the search region (as shown by early-fusion designs like OSTrack~\cite{ye2022joint}), we deliberately chose not to do so due to computational and inference efficiency concerns. The diffusion model already integrates the template and text semantics within the latent space and produces a rich fused representation. Decoding it into an image and then re-encoding it back into features via the backbone is both redundant and computationally expensive. Moreover, such a design would force the pipeline into serial execution—first generation, then tracking—thereby significantly limiting inference speed. To avoid this bottleneck, we directly use the fused features from the diffusion model and perform cross-attention fusion with the search region features after the backbone, balancing effectiveness and efficiency.

\subsection{Multi-Modal Decoder}
\label{MDM}

To solve the misalignment between features from Stable Diffusion U-Net and the image encoder, we propose a novel module for feature alignment and information fusion, which we name Multi-Modal Decoder. This module uses the multi-modal features fused within U-Net to assist the tracking process. Each Multi-Modal Decoder comprises one Attention Pooling Module and one Feature Decoding Module. There are $N$ Multi-Modal Decoders cascaded to sequentially exploit features from U-Net, where $N$ equals the number of U-Net features. As shown in Figure~\ref{fig:stable diffusion u-net} and Figure~\ref{fig:multi-modal decoder}, GLAD generally sets $N$ to 3 for the balance of performance and efficiency.

\subsubsection{Attention Pooling Module}

\begin{figure}[htbp]
\centering
\includegraphics[width=\columnwidth]{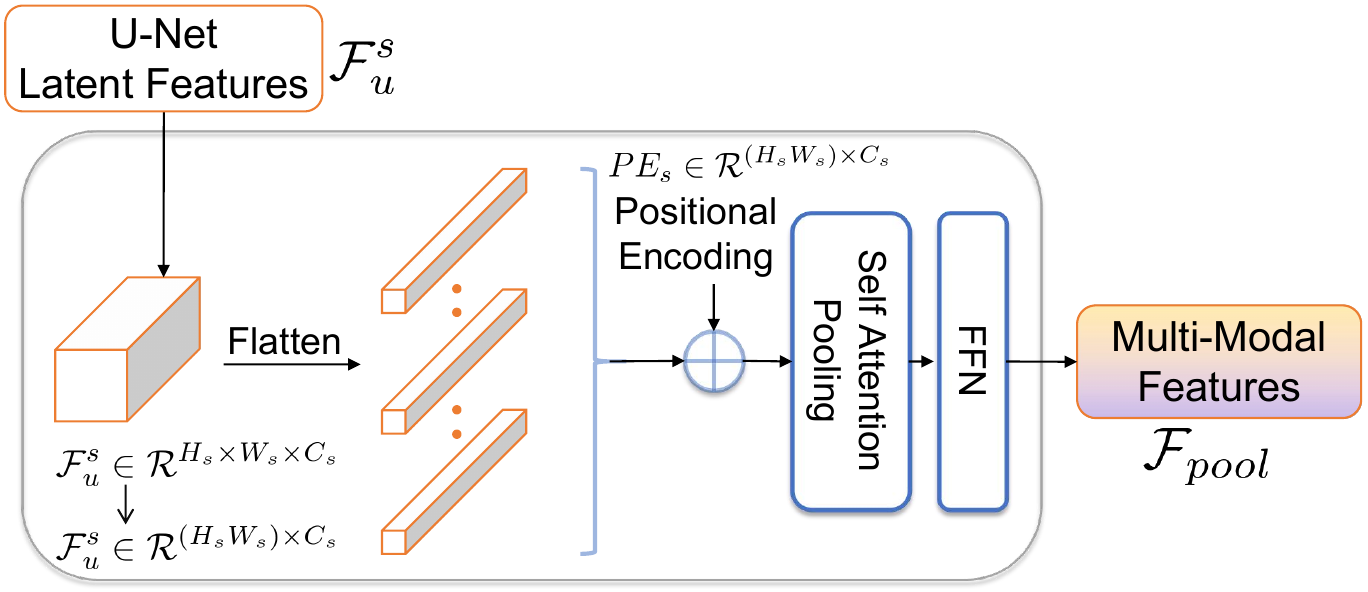}
\caption{Overview of the proposed Attention Pooling Module. It aligns the spatial and dimensional discrepancies between U-Net and visual encoder features while preserving semantic richness for effective multi-modal fusion.}
\label{fig:Pooling}
\end{figure}

Because the multi-modal features generated by the U-Net are inherently incompatible with the visual features extracted by the image encoder in terms of both spatial resolution and feature dimensionality, direct feature interaction becomes infeasible. To resolve this mismatch and simultaneously strengthen the semantic expressiveness of the fused representation, we propose an attention-based pooling module that performs dimension alignment while preserving semantic richness. As illustrated in Figure~\ref{fig:Pooling}, features from the $s$-th Transformer submodule of the U-Net are first denoted as

\begin{equation}
\mathcal{F}^s_u \in \mathbb{R}^{H_s \times W_s \times C_s},
\end{equation}

where $H_s$ and $W_s$ represent the spatial height and width of the feature map, and $C_s$ denotes the channel dimension.  
These features are subsequently flattened into a sequential representation for downstream processing,

\begin{equation}
\mathcal{F}^s_u \in \mathbb{R}^{(H_sW_s) \times C_s}.
\end{equation}

Since this flattening operation inevitably discards the original spatial positional information, we introduce a learnable positional encoding to restore location awareness, defined as

\begin{equation}
PE_s \in \mathbb{R}^{(H_sW_s) \times C_s}.
\end{equation}

At this stage, the multi-modal sequence enriched with positional encoding is expressed as

\begin{equation}
\mathcal{F}^s_m = \mathcal{F}^s_u + PE_s,
\end{equation}

which restores the spatial location cues lost during flattening and prepares the features for further refinement.  

This enriched representation \(\mathcal{F}^s_m\) then serves as the input to the subsequent self-attention pooling stage, where the features are not only semantically enhanced but also appropriately scaled and dimensionally aligned. In this way, the processed multi-modal features can be effectively integrated into the feature decoding module, ensuring both representational adequacy and compatibility for cross-modal fusion.

To be more specific, self-attention pooling operation conducts self-attention on the multi-modal features \(\mathcal{F}^s_m\) itself, promoting the fusion of global spatial information and reducing the feature dimensions to be suitable for the size of visual features. Here, we use \(q_m\), \(k_m\), and \(v_m\) as representations for query, key, and value embedding of \(\mathcal{F}^s_m\) in attention calculation. 
The calculation formula is:
\begin{equation}
    Attn_{m} = \text{Softmax}(\frac{q_mk_m^T}{\sqrt{d}})v_m.
\end{equation}

Finally, a Feed Forward Network (FFN) is applied for output:
\begin{equation}
    \mathcal{F}_{pool} = \text{FFN}(Attn_{tp}),
\end{equation}
and the result $\mathcal{F}_{pool}$ is then sent to the Feature Decoding Module for further semantic enhancement.

\subsubsection{Feature Decoding Module}

\begin{figure}[htbp]
\centering
\includegraphics[width=0.99\columnwidth]{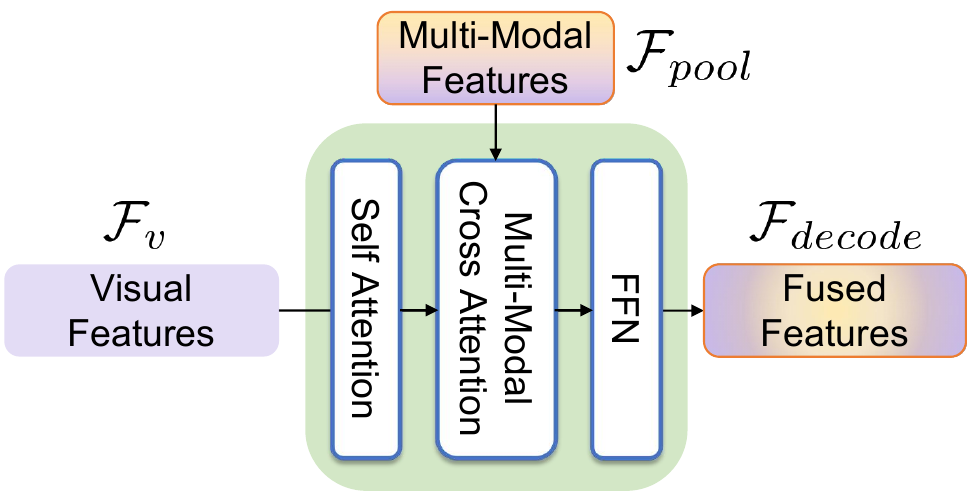}
\caption{Structure of the Feature Decoding Module. 
It enhances the semantic representation and discriminative ability of search features by leveraging multi-modal information from the Attention Pooling Module.
}
\label{fig:Decoder}
\end{figure}

The Attention Pooling Module mainly ensures that the spatial resolution and feature dimensionality are aligned with those of the template features, thereby enabling cross-modal interaction. However, this step alone does not substantially enrich the semantic content of the search features, which remain relatively weak in expressive power. To overcome this limitation, we design a Feature Decoding Module that further leverages the multi-modal features to strengthen their semantic representation and enhance their discriminative capacity for tracking, as illustrated in Figure~\ref{fig:Decoder}.

In the Feature Decoding Module, the visual features are constructed as the concatenation of template features and search features, formally defined as  

\begin{equation}
\mathcal{F}_{tp} = \text{Enc}(I_{tp}), \quad 
\mathcal{F}_{sr} = \text{Enc}(I_{sr}),
\end{equation}

where \(I_{tp}\) and \(I_{sr}\) denote the template image and the search region image, respectively, and \(\text{Enc}(\cdot)\) represents the Image Encoder~\citep{ye2022joint}.  
The combined visual feature representation is then given by  

\begin{equation}
\mathcal{F}_v = \text{Concat}(\mathcal{F}_{tp}, \mathcal{F}_{sr}).
\end{equation}

This concatenated representation allows the model to jointly consider information from both the template and the search region within a unified feature space. To capture long-range dependencies and global spatial relationships, we apply a self-attention mechanism over the visual features \(\mathcal{F}_v\). Specifically, if we denote the concatenated embeddings as query, key, and value matrices \(q_v\), \(k_v\), and \(v_v\), the output of the self-attention layer is computed as  

\begin{equation}
  Output_{SA} = \text{Softmax}\left(\frac{q_vk_v^T}{\sqrt{d}}\right)v_v,
\end{equation}

where \(d\) denotes the scaling factor corresponding to the feature dimension.  
This operation enables the model to adaptively reweight feature responses, ensuring that salient spatial patterns across the template and search regions are effectively emphasized in the decoding process.

Then, the multi-modal features from the Attention Pooling Module, denoted as \(\mathcal{F}_{pool}\), are introduced into the cross-attention operation. Specifically, \(\mathcal{F}_{pool}\) is linearly projected to obtain the key and value matrices:

\begin{equation}
k_{pool} = W_k \mathcal{F}_{pool}, \quad v_{pool} = W_v \mathcal{F}_{pool},
\end{equation}

where \(W_k\) and \(W_v\) are learnable projection matrices.  
Meanwhile, the query embeddings are derived from the output of the self-attention layer, expressed as

\begin{equation}
q_{v} = W_q Output_{SA},
\end{equation}

with \(W_q\) representing the projection matrix for query embeddings.  
The cross-attention output is then calculated as

\begin{equation}
Output_{CA} = \text{Softmax}\left(\frac{q_v k_{pool}^T}{\sqrt{d}}\right) v_{pool},
\end{equation}

where \(d\) denotes the scaling factor corresponding to the feature dimension.  
Finally, a Feed-Forward Network (FFN) is applied channel-wise to the attended features for consolidation, yielding the decoded representation

\begin{equation}
\mathcal{F}_{decode} = \text{FFN}(Output_{CA}),
\end{equation}

which is subsequently forwarded to the Head Network for target location prediction.

\subsection{Head}
\label{Hn}

The head module is responsible for predicting the key points of target bounding boxes. In prior works, this component is often implemented either through a Multi-Layer Perceptron (MLP) or a corner-based fully Convolutional Neural Network (CNN). However, visualization analyses from MixFormer~\citep{cui2022mixformer} reveal that corner-based designs negatively impact the capability of relation modeling modules, as they force the network to attend disproportionately to local boundaries rather than capturing holistic object cues. To mitigate this limitation, we adopt a center-based design, which encourages the network to concentrate on the central position of the target and improves the robustness of bounding box regression.  

Specifically, we employ a fully convolutional center-based localization head~\citep{CenterNet} to estimate the bounding boxes of tracked objects. The head consists of $L$ stacked Conv-BN-ReLU layers, which progressively refine the feature representation for accurate localization. Specifically, the head outputs three types of prediction maps:  

\begin{equation}
C \in [0,1]^{\frac{H_{sr}}{P} \times \frac{W_{sr}}{P}},
\end{equation}

\begin{equation}
O \in [0,1]^{2 \times \frac{H_{sr}}{P} \times \frac{W_{sr}}{P}},
\end{equation}

\begin{equation}
S \in [0,1]^{2 \times \frac{H_{sr}}{P} \times \frac{W_{sr}}{P}},
\end{equation}

where $C$ denotes the classification score map, $O$ represents the local offset map, and $S$ indicates the normalized bounding box size map. Here, $H_{sr}$ and $W_{sr}$ refer to the spatial resolution of the search image, while $P$ is the patch size used in feature extraction.  

Finally, the position corresponding to the maximum classification score in $C$ is selected as the predicted target center. The target bounding box is then obtained by combining this position with the offset and size information provided by $O$ and $S$, respectively. Formally, if $(x_c, y_c)$ denotes the predicted center location, the bounding box can be expressed as  

\begin{equation}
B = (x_c + O_x, \; y_c + O_y, \; S_w, \; S_h),
\end{equation}

where $(O_x, O_y)$ are the local offsets from the offset map $O$, and $(S_w, S_h)$ are the predicted width and height of the bounding box from the size map $S$. 

\begin{table*}[htbp]
    \centering
    \caption{State-of-the-art Comparison on LaSOT, LaSOT$_{\text{ext}}$, TNL2K and OTB99-lang. The best two results are in \textcolor{red}{\textbf{red}} and \textcolor{blue}{\textbf{blue}} fonts.}
    \renewcommand{\arraystretch}{1.2} 
    \resizebox{\linewidth}{!}{
        \begin{tabular}{c | ccc | cc | ccc | cc}
        \toprule
        \multirow{2}{*}{Method} & \multicolumn{3}{c|}{LaSOT} & \multicolumn{2}{c|}{LaSOT$_{\text{ext}}$} & \multicolumn{3}{c|}{TNL2K} & \multicolumn{2}{c}{OTB99-lang} \\
        \cmidrule{2-11}
         & $AUC(\%)$ & $P_{Norm}(\%)$ & $P(\%)$ & $AUC(\%)$ & $P(\%)$ & $AUC(\%)$ & $P_{Norm}(\%)$ & $P(\%)$ & $AUC(\%)$ & $P(\%)$  \\
        \midrule
        \multicolumn{10}{l}{\quad \textit{Visual tracking method (vision-only)}} \\
        \midrule
        DiMP~\citep{bhat2019learning} & 56.9 & 65.0 & 56.7 &  39.2 & 45.1 & 44.7 & 51.3 & 43.4 & - & - \\
        TransT~\citep{chen2021transformer} & 64.9 & 73.8 & 69.0 & 45.1 & 51.2 & 50.7 & 57.1 & 51.7 & - & - \\
        MixViT~\citep{cui2022mixformer} & 69.6 & 79.9 & 75.9 & - & - & - & - & - & - & - \\
        MixFormerV2~\citep{cui2023mixformerv2} & 70.6 & 80.8 & 76.2 & 50.6 & 56.9 & 57.4 & - & 58.4 & - & - \\ 
        OSTrack~\citep{ye2022joint} & 69.1 & 78.7 & 75.2 & 47.4 & 53.3 & 54.3 & - & - & - & - \\
        SeqTrack-B256~\citep{DBLP:conf/cvpr/ChenPWLH23} & 69.9 & 79.7 & 76.3 & 49.5 & 56.3 & 54.9 & - & - & - & - \\
        SeqTrack-L384~\citep{DBLP:conf/cvpr/ChenPWLH23} & 72.5 & 81.5 & 79.3 & 50.7 & 57.5 & 57.8 & - & - & - & - \\
        SUTrack-L256~\citep{chen2025sutrack} & 73.5 & 83.3 & 80.9 & 54.0 & 61.7 & - & - & - & - & - \\
        SUTrack-L256~\citep{chen2025sutrack} & 75.2 & 84.9 & 83.2 & 53.6 & 60.5 & - & - & - & - & - \\
        ARTrack-B256~\citep{wei2023autoregressive} & 70.4 & 79.5 & 76.6 & 46.4 & 52.3 & 57.5 & - & - & - & - \\
        ARTrack-L384~\citep{wei2023autoregressive} & 73.1 & 82.2 & 80.3 & 52.8 & 59.7 & 60.3 & - & - & - & - \\
        \midrule  
        \multicolumn{10}{l}{\quad \textit{Vision-language tracking method}} \\
        \midrule
        RTTNLD~\citep{feng2020real} & 35.0 & - & 35.0 & - & - & 25.0 & 33.0 & 27.0 & 61.0 & 79.0 \\
        SNLT~\citep{feng2021siamese} & 54.0 & 63.6 & 57.4 & 26.2 & 30.0 & 27.6 & 35.9 & 41.9 & 66.6 & 84.8 \\
        TNL2K-\MakeUppercase{\romannumeral 2}~\citep{wang2021towards} & 51.3 & - & 55.4 & - & - & 42.0 & 50.0 & 42.0 & 68.0 & 88.0 \\
        GTI~\citep{yang2020grounding} & 47.8 & - & 47.6 & - & - & - & - & - & 58.1 & 73.2 \\
        Li et al.~\citep{li2022cross} & 53.0 & 56.0 & - & - & - & 44.0 & 52.0 & 45.0 & 69.0 & 91.0 \\
        $\text{VLT}_{\text{TT}}$~\citep{guo2022divert} & 67.3 & - & 72.1 & 48.4 & 55.9 & 53.1 & - & 53.3 & {\color{red} \textbf{76.4}} & {\color{blue} \textbf{93.1}} \\
        TransVLT~\citep{zhao2023transformer} & 66.4 & - & 70.8 & - & - & 56.0 & 61.7 & - & 69.9 & 90.6 \\
        JointNLT~\citep{zhou2023joint} & 60.4 & 69.4 & 63.6 & - & - & 56.9 & 73.6 & 58.1 & 65.3 & 85.6 \\
        QueryNLT~\citep{shao2024context} & 59.9 & 69.6 & 63.5 & - & - & 57.8 & 75.6 & 58.7 & 66.7 & 88.2 \\
        MMTrack~\citep{zheng2023toward} & 70.0 & 82.3 & 75.7 & 49.4 & 55.3 & 58.6 & 75.2 & 59.4 & 70.5 & 91.8 \\
        DecoupleTNL~\citep{ma2023tracking} & 64.9 & - & 67.1 & - & - & 40.7 & - & 40.0 & 69.5 & 92.8 \\
        All-in-one~\citep{zhang2023all} & {\color{blue} \textbf{71.7}} & {\color{blue} \textbf{82.4}} & {\color{blue} \textbf{78.5}} & {\color{blue} \textbf{54.5}} & {\color{blue} \textbf{62.0}} & 55.3 & - & 57.2 & 71.0 & 93.0 \\ 
        ChatTracker~\citep{DBLP:conf/nips/SunYCZHLLW24} & {\color{blue} \textbf{71.7}} & 80.9 & 77.5 & - & - & 59.6 & 76.3 & 62.1 & - & - \\
        DMTrack-256~\citep{DBLP:conf/ijcai/ZhangZLMS24} & 66.8 & - & 72.7 & 47.3 & 52.1 & 57.7 & - & 59.9 & 69.3 & 90.9 \\
        DMTrack-384~\citep{DBLP:conf/ijcai/ZhangZLMS24} & 70.1 & - & 76.5 & 51.1 & 58.3 & {\color{blue} \textbf{61.6}} & - & {\color{blue} \textbf{65.4}} & 71.2 & 92.3 \\
        \midrule
        GLAD-B256 (Ours) & 69.5 & 79.6 & 74.2 & 50.6 & 55.7 & 59.7 & {\color{blue} \textbf{77.6}} & 62.2 & 69.6 & 92.4 \\
        GLAD-L384 (Ours) & {\color{red} \textbf{76.0}} & {\color{red} \textbf{86.8}} & {\color{red} \textbf{85.1}} & {\color{red} \textbf{55.9}} & {\color{red} \textbf{64.6}} & {\color{red} \textbf{62.1}} & {\color{red} \textbf{80.4}} & {\color{red} \textbf{66.2}} & {\color{blue} \textbf{71.4}} & {\color{red} \textbf{93.2}}  \\
        \bottomrule
        \end{tabular}
    }
    \label{tab:GLAD_SOTA_Comparison}
\end{table*}

\subsection{Optimization for Efficiency}
\label{OFE}

The native Stable Diffusion\footnote{Stable Diffusion v1.5 is available at \href{https://huggingface.co/stable-diffusion-v1-5/stable-diffusion-v1-5}{huggingface - stable-diffusion-v1-5}, a mirror of the deprecated ruwnayml repo.} model encounters problems with inference efficiency in image generation, mainly resulting from a large number of denoising steps. 
It poses significantly negative impacts on training and inference for the proposed GLAD method. To solve this problem, GLAD applies a technique that can accelerate the denoising speed of diffusion models, called Latent Consistency Model (LCM)~\citep{luo2023lcm}, to optimize the efficiency of the generative multi-modal fusion process.

Inspired by Consistency Models (CM)~\citep{song2023consistency}, this technique integrates the principles of Latent Diffusion Models (LDM)~\citep{rombach2022high} to design a Latent Consistency Model (LCM). It employs a distillation-based approach to finetune diffusion models, yielding a training-free acceleration plugin named LCM-LoRA~\citep{luo2023lcm}. By combining the Stable Diffusion v1.5 model with the LCM-LoRA model, the denoising steps are reduced from the original 20 to 50 steps to only 2 to 8 steps, greatly enhancing the speed of the generative multi-modal fusion process and finally promoting efficiency in training and inference.

Through this approach, during inference, the diffusion process is applied only once at the first frame for initialization. Specifically, we adopt the SD-v1.5-inpainting model and perform a single denoising step, after which the fused features obtained are fixed and reused for subsequent tracking. In practical applications, the diffusion preparation time is relatively short—roughly comparable to the cost of a single-frame forward pass—making its impact on the overall inference latency negligible.

\section{Experiment}\label{sec: experiment}

\subsection{Experimental Settings}

GLAD is implemented using Python 3.8.18, Diffusers 0.23.0, and PyTorch 2.1.1. The training process is conducted on 4 Nvidia Quadro RTX A6000 GPUs, while we use 1 Nvidia GeForce RTX 3090 GPU for inference and efficiency test. 
For the pretrained backbone, we adopt the widely used Stable Diffusion v1.5 model as the foundation for generative multi-modal fusion, owing to its strong text-to-image alignment capability and robust denoising performance. In addition, we employ the MAE-pretrained ViT~\citep{DBLP:conf/cvpr/HeCXLDG22} as the image encoder to provide high-quality visual representations. The Stable Diffusion v1.5 model contains the Text Encoder, the Variational Autoencoder (VAE), the Forward process, and the U-Net, which collectively support multi-modal representation learning. On top of this architecture, we design three attention pooling modules and three feature decoding modules in the Multi-Modal Decoder, both of which are initialized with the Xavier scheme~\citep{glorot2010understanding} to stabilize training and accelerate convergence.

For GLAD-Base, the input resolution of the search image is set to 
$256 \times 256$, while for GLAD-Large, the resolution is increased to 
$384 \times 384$, providing higher spatial fidelity for more challenging tracking 
scenarios. Similarly, the template image is configured with a resolution of 
$128 \times 128$ in GLAD-Base and $192 \times 192$ in GLAD-Large, ensuring a 
balanced trade-off between computational efficiency and representational quality. 
In order to align with the requirements of the Stable Diffusion v1.5 backbone, 
the template images from both variants are further reshaped to $512 \times 512$ 
during the template inpainting stage, which guarantees consistency with the 
generative module and preserves the semantic richness of reconstructed features.

For the training stage, we include the training splits of GOT-10k~\citep{huang2019got}, LaSOT~\citep{fan2019lasot}, OTB99-lang~\citep{li2017tracking}, TNL2K~\citep{wang2021towards}, TrackingNet\citep{muller2018trackingnet}, RefCOCOg~\citep{DBLP:conf/cvpr/MaoHTCY016} to train our model. 
The whole training process takes 300 epochs, using AdamW\citep{loshchilov2017decoupled} optimizer with weight decay of \(10^{-4}\) and gradient clipping of 0.1. Besides, we employ a compound loss function that combines L1 loss, to enforce precise coordinate regression, and GIoU loss as follows, to enhance spatial alignment and robustness in bounding box prediction.
\begin{equation}
    L_{loc} = \lambda_{L1} L_1(B_i, \hat{B_i}) + \lambda_{giou} L_{giou}(B_i, \hat{B_i}),
\end{equation}
where $\lambda_{L1}=5$ and $\lambda_{giou}=2$ are the weights of the two losses, $B_i$ is the ground-truth bounding box and $\hat{B_i}$ is the predicted bounding box of the targets. 

During training, the warm-up strategy is adopted. The learning rate first increases to \(4\times 10^{-4}\) linearly at the first 30 epochs and then drops to \(4\times 10^{-5}\) after 240 epochs. 
For GLAD, pretraining is necessary for Image Encoder. We pretrain the Image Encoder with the Visual Object Tracking task to get an appropriate initialization for the Vision-Language Tracking task.
Moreover, we adopt the DeepSpeed\citep{rasley2020deepspeed} ZeRO\citep{rajbhandari2020zero} Stage 2 strategy during training. This strategy partitions the optimizer states and gradients into parallel workflows for each device, significantly reducing the GPU memory consumption during training and improving the training speed.

While for the evaluate stage, we evaluate our model for vision-language tracking on the mainstream tracking benchmarks with natural language descriptions, including LaSOT~\citep{fan2019lasot}, TNL2K~\citep{wang2021towards}, OTB99~\citep{li2017tracking}
and LaSOT$_{\text{ext}}$~\citep{fan2021lasot}. 
Besides, we also evaluate our model on the traditional visual tracking dataset GOT-10k~\citep{huang2019got} to verify the tracking performance in a vision-only scenario. 
We generally follow the convention~\citep{chen2021transformer, ye2022joint, zhang2020ocean} by applying a simple Hanning window penalty to the output of the head network to utilize positional prior information.
For acceleration techniques, we only used FP16 precision and Flash Attention, both of which are commonly adopted acceleration strategies.

\subsection{Comparison with the State-of-the-art Methods}

This section demonstrates the performance of GLAD on various visual language tracking datasets and compares it with current mainstream methods, which is shown in Table~\ref{tab:GLAD_SOTA_Comparison}.
Compared to vision-language trackers,
GLAD-L384 establishes a new SOTA on LaSOT~\citep{fan2019lasot}, LaSOT$_{\text{ext}}$~\citep{fan2021lasot} and TNL2K~\citep{wang2021towards}, significantly exceeds previous methods such as All-in-one~\citep{zhang2023all}, DMTrack-384~\citep{DBLP:conf/ijcai/ZhangZLMS24}, JointNLT~\citep{zhou2023joint} and QueryNLT~\citep{shao2024context}.

\noindent\textbf{LaSOT \& LaSOT\(_{ext}\).} 
They are extensions of traditional visual tracking benchmarks \citep{fan2019lasot,li2017tracking} by adding text labels, focusing on long-term tracking challenges. 
Furthermore, LaSOT\(_{ext}\) also includes many similar distractors, further complicating the tracking task.

As shown in Table~\ref{tab:GLAD_SOTA_Comparison}, GLAD-L384 achieved the highest score in the $AUC$ metric, surpassing All-in-one~\citep{zhang2023all} by 4.3\%, indicating the effectness of our generative fusion paradigm on making full use of high-semantic texts for improvement of target locating.
Compared to pure vision models like SeqTrack~\citep{DBLP:conf/cvpr/ChenPWLH23} and ARTrack~\citep{wei2023autoregressive}, GLAD-L384 also achieves better performance, which further validates the effectiveness of the template inpainting approach employed by our method.

\begin{table*}[htbp]
\centering
\setlength{\tabcolsep}{16pt}
\caption{Params and MACs analysis of GLAD. All tests were conducted on a RTX 3090 GPU.}
\begin{tabular}{c|c|c|c}
\toprule
     Model & Module & Params(M) & MACs(G) \\
     \midrule
     \multirow{5}{*}{GLAD-B256} & Image Encoder & 85.6 & 27.4 \\
    & Pooling & 17.7 & 3.4 \\
    & Decoding & 28.3 & 13.4 \\
    & Head & 6.5 & 1.7 \\
    \cmidrule{2-4}
    & Total & 138.2 & 41.1 \\
    \midrule
    \multirow{5}{*}{GLAD-L384} & Image Encoder & 303.1 & 218.1 \\
    & Pooling & 18.7 & 3.6 \\
    & Decoding & 50.4 & 52.0 \\
    & Head & 8.2 & 4.7 \\
    \cmidrule{2-4}
    & Total & 380.4 & 259.4 \\
    \midrule
    JointNLT & Total & 153.0 & 42.0 \\
    \midrule
    MMTrack & Total & 176.9 & - \\
    \bottomrule
\end{tabular}
\label{tab:macs}
\end{table*}

\noindent\textbf{TNL2K.} 
It is a large-scale benchmark specifically constructed for natural language-guided tracking, consisting of over two thousand video sequences with diverse target categories and fine-grained linguistic annotations. 
The dataset enables evaluation of the interaction between visual and textual modalities, making it a common choice for studying language-assisted tracking \citep{wang2021towards}.  

As shown in Table~\ref{tab:GLAD_SOTA_Comparison}, GLAD-L384 outperforms the recent DMTrack-384 by 0.5\% in AUC and 0.8\% in P. 
Compared to  ChatTracker~\citep{DBLP:conf/nips/SunYCZHLLW24} which employs multimodal large language models to generate high-quality text annotations for addressing static text limitations, our textual target-context guidance module achieves superior performance by 2.5\% in AUC and 4.1\% in P.

\noindent\textbf{OTB99.} 
It is a classical short-term tracking benchmark containing 99 video sequences with manually annotated bounding boxes across a variety of common tracking scenarios. 
An additional extension, OTB-sentences, provides one natural language description for each sequence, allowing the benchmark to be used in vision-language tracking studies as well \citep{li2017tracking}.  

As shown in Table~\ref{tab:GLAD_SOTA_Comparison}, GLAD-L384 also exhibits leading performancee in the $AUC$ metric, surpassing All-in-one~\citep{zhang2023all} by 0.4\% in AUC and 0.2\% in P. Compared to JointNLT~\citep{zhou2023joint} which uses joint visual
grounding and tracking framework, we still achieves superior performance by 4.7\% in AUC and 5.0\% in P,
demonstrating the effectiveness of language information and GLAD's strong capability in utilizing text features.

\begin{table}[htbp]
\centering
\caption{Efficiency analysis of GLAD. All tests were conducted on a RTX 3090 GPU.}
\renewcommand{\arraystretch}{1.2} 
\begin{tabular}{c | c | c}
    \toprule
    Method & Speed (FPS) & LaSOT AUC(\%) \\
    \midrule
    JointNLT & 39 & 60.4 \\
    TransVLT & 36 & 66.4 \\
    VLT$_{\text{TT}}$ & 48 & 67.3 \\
    MMTrack & 45 & 70.0 \\
    \midrule
    \textbf{GLAD-B256} & \textbf{65} & \textbf{69.5} \\
    \textbf{GLAD-L384} & \textbf{44} & \textbf{76.0} \\
    \bottomrule
\end{tabular}
\label{tab:GLAD_efficiency}
\end{table}

\noindent\textbf{Inference Speed} 
It is a critical factor for practical visual tracking applications, as many scenarios such as autonomous driving, robotics, and surveillance require both accurate and fast target localization. High inference speed ensures that the tracker can respond promptly to dynamic changes in the environment, making it suitable for time-sensitive tasks and large-scale deployment.

To evaluate the efficiency of our method, we conduct a detailed statistical analysis of tracking speed and computational cost.
For computational cost, we reported the parameter count of each module and make a comparison, as shown in Table~\ref{tab:macs}.
For tracking speed, we compare GLAD-L384 with several representative approaches in Table~\ref{tab:GLAD_efficiency}. The results indicate that GLAD-L384 runs at 44 FPS, which is comparable to $\text{VLT}_{\text{TT}}$~\citep{guo2022divert} and MMTrack~\citep{zheng2023toward}. Notably, this high inference speed is achieved while still maintaining competitive accuracy, demonstrating that our generative multi-modal fusion framework effectively balances computational efficiency and tracking performance for real-world applications.

\begin{table*}[htbp]
\centering
\setlength{\tabcolsep}{16pt}
\caption{State-of-the-art comparison on the vision-only dataset GOT-10k. The best two results are shown in \textcolor{red}{\textbf{red}} and \textcolor{blue}{\textbf{blue}} fonts.}
\renewcommand{\arraystretch}{1.2} 
\begin{tabular}{c | c | c | c}
    \toprule
    Method & $AO(\%)$ & $SR_{0.5}(\%)$ & $SR_{0.75}(\%)$ \\
    \midrule
    \multicolumn{4}{l}{\quad \textit{Visual tracking method (vision-only)}} \\
    \midrule
    TransT~\citep{chen2021transformer} & 67.1 & 76.8 & 60.9\\
    STARK~\citep{STARK} & 68.8 & 78.1 & 64.1 \\ 
    MixFormer~\citep{cui2022mixformer} & 71.2 & 79.9 & 65.8 \\
    MixViT ~\citep{cui2022mixformer} & 72.7 & 82.3 & 70.8 \\
    OSTrack-B256~\citep{ye2022joint} & 71.0 & 80.4 & 68.2 \\
    OSTrack-L384~\citep{ye2022joint} & 73.7 & 83.2 & \color{blue}\textbf{70.8} \\
    AiATrack~\citep{AiATrack} & 69.6 & 80.0 & 63.2 \\
    SimTrack~\citep{SimTrack} & 68.6 & 78.9 & 62.4\\
    \midrule
    \multicolumn{4}{l}{\quad \textit{Vision-language tracking method}} \\
    \midrule
    RomTrack~\citep{Cai_2023_ICCV} & 72.9 & 82.9 & 70.2 \\
    VLT$_{\text{SCAR}}$~\citep{guo2022divert} & 61.4 & 72.4 & 52.3 \\
    VLT$_{\text{TT}}$~\citep{guo2022divert} & 69.4 & 81.1 & 64.5 \\
    \midrule
    GLAD-B256(Ours) & \color{blue}\textbf{73.0} & \color{blue}\textbf{83.4} & 69.8 \\
    GLAD-L384(Ours) & \color{red}\textbf{74.5} & \color{red}\textbf{84.7} & \color{red}\textbf{73.1} \\
    \bottomrule
\end{tabular}
\label{tab:GLAD_GOT-10k_Comparison}
\end{table*}

\subsection{Extra comparison on traditional visual tracking dataset}

Besides, we also evaluate our model on the traditional visual tracking dataset GOT-10k~\citep{huang2019got} to verify the tracking performance in a vision-only scenario.
Since GOT-10k contains only visual information without any textual descriptions, it provides an ideal setting to examine whether our generative multimodal fusion paradigm can still contribute to performance improvement when the model relies solely on visual cues.
This evaluation also helps determine whether the effectiveness of GLAD stems from its generative modeling ability itself rather than merely from additional language guidance.

As summarized in Table~\ref{tab:GLAD_GOT-10k_Comparison}.
Our GLAD surpasses VLT$_{\text{TT}}$~\citep{guo2022divert} by 3.6\% on the AO metric and achieves performance comparable to state-of-the-art vision-only trackers such as MixFormer~\citep{cui2022mixformer} and OSTrack~\citep{ye2022joint}.
This demonstrates that even in the absence of textual input, GLAD maintains strong tracking capability and robust visual understanding.
These results further validate that the proposed generative paradigm enables the extraction of semantically richer and more discriminative visual representations, ultimately enhancing tracking robustness across different settings.

\subsection{Ablation Experiment}

To thoroughly evaluate the effectiveness of our proposed method, we conduct comprehensive ablation studies on each major module within the framework. The experiments are carried out on two widely recognized vision-language tracking benchmarks, LaSOT~\citep{fan2019lasot} and TNL2K~\citep{wang2021towards}, which together cover diverse tracking scenarios and varying degrees of textual-visual correspondence. All experiments are performed under the same training and inference configurations to ensure fairness and consistency in comparison, allowing us to clearly assess the contribution and robustness of each module.

\begin{figure*}[htbp]
    \centering
    \subfloat[Feature Modulation]{\includegraphics[width=0.48\linewidth]{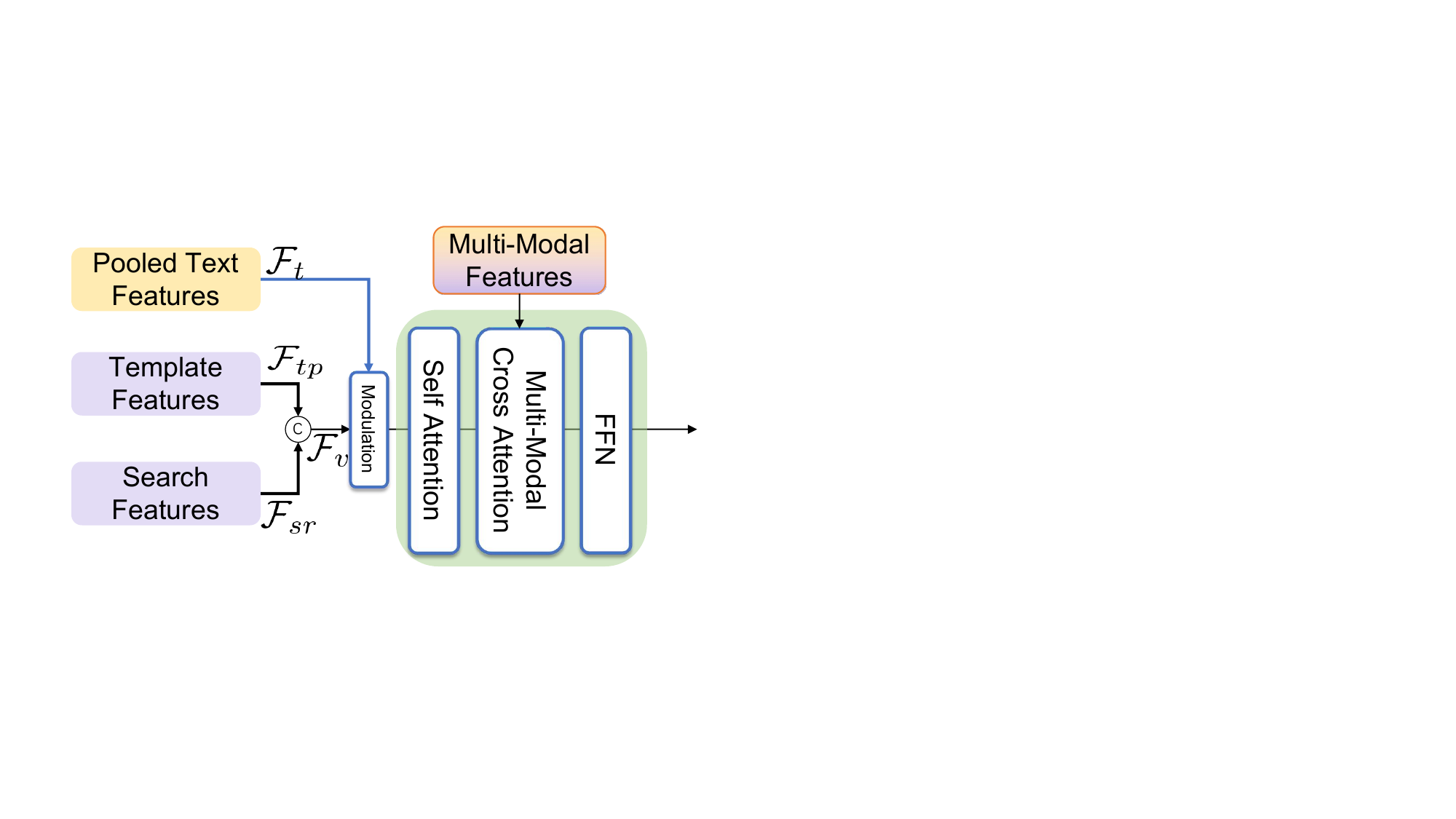}
    \label{fig:feature modulation}}
    \hspace{5pt}
    \subfloat[Feature Concatenation]{\includegraphics[width=0.42\linewidth]{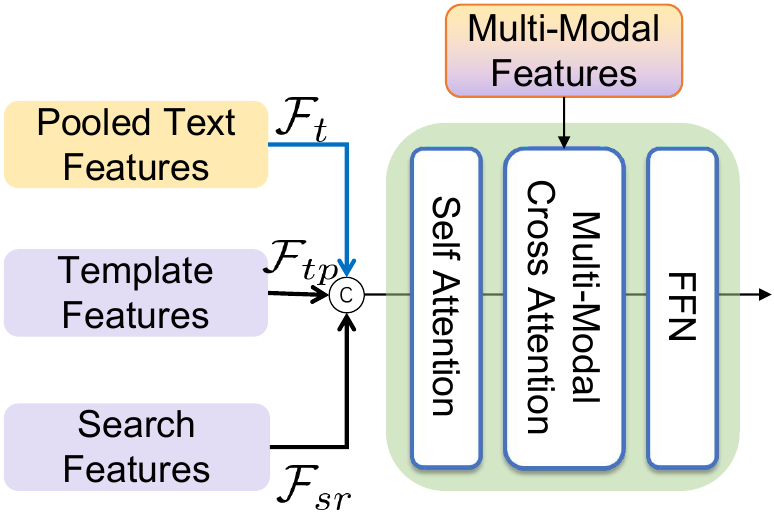}
    \label{fig:feature concatenation}}
    \caption{Two types of text–visual feature fusion methods are illustrated. The template and search features are extracted from the Image Encoder, while the pooled text features are obtained from the Text Encoder in Stable Diffusion, which is equivalent to the Text Encoder used in CLIP. The multi-modal features are produced by the attention pooling module. In the left design, a simple feature modulation is applied to introduce textual information, whereas in the right design, feature concatenation is employed for fusion.}
    \label{fig:decoder_Ablation}
\end{figure*}

\subsubsection{Exploration of Feature Decoding Module}
We observed that in current vision-language tracking datasets, the textual descriptions often fail to align precisely with the visual content of the search images, primarily due to the inherent semantic gap between linguistic and visual modalities during the tracking process. This mismatch leads to inaccurate or noisy guidance when textual features are directly fused with the visual representations of search regions, potentially degrading overall performance. To further verify the impact of this issue, we design two variant models for comparative analysis, each aiming to isolate and evaluate the role of textual-visual interaction under different fusion strategies.

\noindent\textbf{Feature Modulation: }
As shown in the left part of Figure~\ref{fig:feature modulation}, the template features \(\mathcal{F}_{tp}\) and search features \(\mathcal{F}_{sr}\) are first concatenated into the combined visual representation \(\mathcal{F}_v\), which is then processed by a simple feature modulation block before being fed into the decoder. This variant aims to inject linguistic cues into the visual pathway through a lightweight modulation mechanism while preserving the original spatial structure of the visual features. The modulation is performed between the visual features \(\mathcal{F}_v\) and the textual features \(\mathcal{F}_{t}\) extracted from the text encoder, formulated as:
\begin{equation}
\mathcal{F}_{mod} = \mathcal{F}_v \odot \text{Linear}(\mathcal{F}_{t}) + \mathcal{F}_v,
\end{equation}
where \(\odot\) denotes the Hadamard product and \(\text{Linear}(\cdot)\) represents a learnable linear projection for dimensional alignment. 

As shown in Table~\ref{tab:ablation}, results of Exp-Decode-M show a decrease \(\delta_1\) of 0.8\% / 0.6\% in AUC and 1.1\% / 0.7 \% in P on LaSOT / TNL2k separately. 

\noindent\textbf{Feature Concatenation: }As shown in the right part of Figure~\ref{fig:feature concatenation}, another variant directly concatenates the template features \(\mathcal{F}_{tp}\) and the search features \(\mathcal{F}_{sr}\) with the textual features \(\mathcal{F}_{t}\) generated from the text encoder in the generative diffusion module. This design aims to achieve a more explicit multimodal integration by merging all modalities into a unified representation before decoding. The concatenation operation can be formulated as:
\begin{equation}
\mathcal{F}_{cat} = \text{Concat}(\mathcal{F}_{tp}, \mathcal{F}_{sr}, \mathcal{F}_{t}),
\end{equation}
where \(\text{Concat}(\cdot)\) denotes channel-wise concatenation along the feature dimension. The fused features are subsequently fed into the decoder for joint reasoning and target localization. 

As outlined in Table~\ref{tab:ablation}, the results of Exp-Decode-C show a decrease \(\delta_2\) of 1.1\% / 1.0\% in AUC and 1.6 \% / 1.2 \% in P on LaSOT / TNL2K respectively. 

From the two experiments discussed above, it can be concluded that directly incorporating textual features into the decoding process leads to suboptimal results, as the semantic information from language cannot be properly aligned with the visual representation of the search region. This mismatch introduces noise rather than providing meaningful guidance for target localization.

\begin{table*}
    \caption{Exploration studies. $\delta_1$, $\delta_2$, …, and $\delta_8$ denote the numerical differences compared with GLAD-B256.}
    \renewcommand{\arraystretch}{1.2} 
    \centering
    \begin{tabular}{ c | cc | cc }
        \toprule
        \multirow{2}{*}{\bf Method} & \multicolumn{2}{c|}{LaSOT} & \multicolumn{2}{c}{TNL2K} \\ 
        \cmidrule{2-5}
        & $AUC(\%)$ & $P(\%)$ & $AUC(\%)$ & $P(\%)$ \\
        \midrule
        GLAD-B256 & 69.5 & 74.2 & 59.7 & 62.2 \\
        \midrule
        Exp-Decode-M \((\delta_1)\) & 68.7 (\underline{-0.8}) & 73.1 (\underline{-1.1}) & 59.1 (\underline{-0.6}) & 61.5 (\underline{-0.7}) \\
        Exp-Decode-C \((\delta_2)\) & 68.4 (\underline{-1.1}) & 72.6 (\underline{-1.6}) & 58.7 (\underline{-1.0}) & 61.0 (\underline{-1.2}) \\
        Exp-UNet-5th \((\delta_3)\) & 66.5 (\underline{-3.0}) & 68.3 (\underline{-5.9}) & 56.4 (\underline{-3.3}) & 58.7 (\underline{-3.5}) \\
        Exp-UNet-6th \((\delta_4)\) & 68.6 (\underline{-0.9}) & 73.2 (\underline{-1.0}) & 59.3 (\underline{-0.4}) & 61.1 (\underline{-1.1}) \\
        Exp-UNet-7th \((\delta_5)\) & 67.8 (\underline{-1.7}) & 70.4 (\underline{-3.8}) & 58.5 (\underline{-1.2}) & 60.9 (\underline{-1.3}) \\
        Exp-UNet-10th \((\delta_6)\) & 69.9 (\underline{+0.4}) & 75.1 (\underline{+0.9}) & 60.2 (\underline{+0.5}) & 63.5 (\underline{+1.3}) \\
        Exp-Transformer \((\delta_7)\) & 67.2 (\underline{-2.3}) & 69.7 (\underline{-4.5}) & 58.3 (\underline{-1.4}) & 61.1 (\underline{-1.1}) \\
        Exp-SDv3 \((\delta_8)\) & 70.2 (\underline{+0.7}) & 76.2 (\underline{+2.0}) & 60.9 (\underline{+1.2}) & 62.5 (\underline{+0.3}) \\
        \bottomrule
    \end{tabular}
    \label{tab:ablation}
\end{table*}

In contrast, leveraging the multi-modal features generated by the proposed Attention Pooling Module enables a more structured and semantically consistent integration of textual cues. Through this mechanism, the text information is effectively embedded into the visual domain, enhancing the quality of the target representation and improving overall tracking performance. These results further verify the effectiveness and superiority of our Attention Pooling Module in bridging the semantic gap between vision and language for robust multimodal tracking.

\subsubsection{Utilization of U-Net Features}
GLAD leverages features extracted from three intermediate layers of the U-Net architecture, specifically the 5-th, 6-th, and 7-th Transformer modules in the contraction path. These submodules correspond to progressively deeper layers responsible for feature extraction, where semantic richness increases while spatial resolution decreases. Therefore, it is reasonable to utilize these layers, as they contain high-level semantic information essential for constructing effective multi-modal representations. 

We further analyze the contribution of different U-Net layers through ablation experiments, as summarized in Table~\ref{tab:ablation}. When removing the 5-th Transformer module (Exp-UNet-5th, $\delta_3$), the performance drops noticeably, with decreases of 3.0\% / 5.9\% in AUC and 3.3\% / 3.5\% in precision on LaSOT / TNL2K, respectively. Removing the 6-th module (Exp-UNet-6th, $\delta_4$) leads to a relatively smaller degradation, resulting in 0.9\% / 1.0\% AUC drops and 0.4\% / 1.1\% precision drops on the two datasets. Similarly, excluding the 7-th module (Exp-UNet-7th, $\delta_5$) causes consistent performance decreases of 1.7\% / 3.8\% in AUC and 1.2\% / 1.3\% in precision. These results indicate that each selected layer contributes positively to the final performance, and removing any single layer inevitably degrades tracking accuracy.

\begin{figure*}[htbp]
\centering
\includegraphics[width=0.9\textwidth]{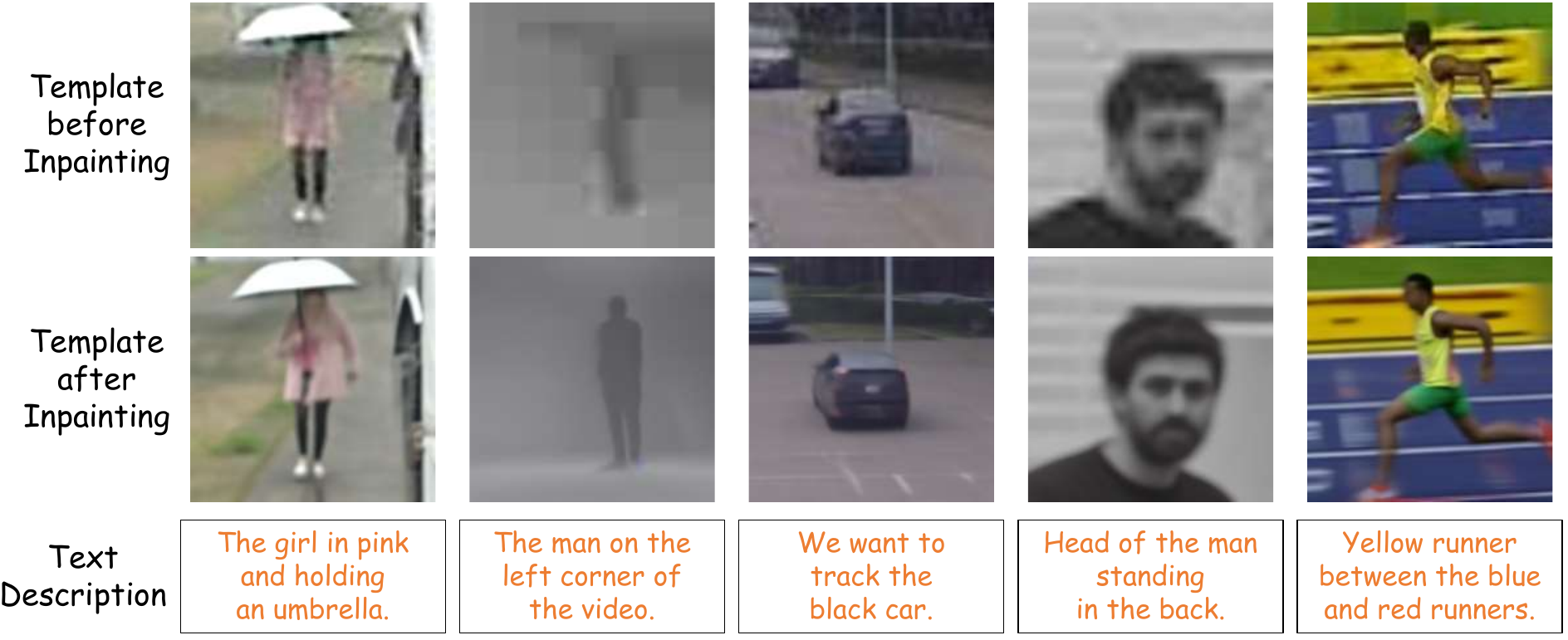}
\caption{Visualizations of Template Inpainting. 
The first row shows templates before inpainting, the second row shows templates after inpainting, and the third row presents the corresponding text descriptions. 
The restored templates exhibit clearer textures and richer semantics, facilitating more accurate target representation.}
\label{fig: vis}
\end{figure*}

On the other hand, we also notice that the submodules along the expansion path correspond to the feature reconstruction and fusion stages, where the feature dimensions gradually become shallower. To further explore the potential of these representations, we investigate whether utilizing features from this stage could provide additional benefits to multi-modal fusion.  

Specifically, we conduct an ablation experiment by additionally introducing the 10-th Transformer module from the expansion path. As shown in Exp-UNet-10th of Table~\ref{tab:ablation}, a slight improvement of $\delta_3 = 0.4\% \,/\, 0.5\%$ in AUC on LaSOT / TNL2k respectively can be observed, indicating that the expansion path still provides useful information for multi-modal feature modeling.  

However, it can also be seen from Figure~\ref{fig:stable diffusion u-net} that after the 7-th Transformer module, the spatial resolution of U-Net features increases significantly. Employing such high-resolution features would substantially raise the computational overhead for both the attention pooling and feature decoding modules. Therefore, despite the potential benefits of expansion path features, they are less suitable for a simple and efficient tracking algorithm like ours, where the trade of inference speed and resource utilization are critical considerations.

In addition, we replace the diffusion (U-Net) module as a whole to further analyze its contribution. Specifically, we retain the original model architecture but remove the generative interaction introduced by diffusion. Instead, the text and template are individually encoded into features and fused using a standard transformer module. The fused features are then fed into the same Multi-Modal Decoder, ensuring consistency with the original pipeline.

 The experimental results are presented in Exp-Transformer of Table~\ref{tab:ablation}. As shown, this variant results in a performance degradation, with AUC drops of 2.3\% / 4.5\% and precision decreases of 1.4\% / 1.1\% on LaSOT / TNL2K, respectively, indicating that the generative interaction is able to better capture informative feature representations.

\subsubsection{Other Diffusion Backbone}

To further demonstrate the effectiveness and robustness of the proposed generative fusion paradigm, we additionally explore the use of alternative diffusion backbones. Specifically, we replace the original Stable Diffusion v1.5 model with Stable Diffusion v3 while keeping the rest of the tracking framework unchanged.As shown in Table~\ref{tab:ablation}, a consistent performance of $\delta_8 = +0.7\% / +2.0\% /$ on LaSOT and $\delta_8 = +1.2\% / +0.3\%$ on TNL2K can be observed across the four benchmarks, indicating that the proposed framework maintains strong tracking performance when adopting a different diffusion backbone.

Besides that, we also believe that fine-tuning the diffusion model on tracking datasets could lead to even better results, as the distribution of the original training data for the diffusion model differs from that of tracking data. However, due to the high computational cost associated with fine-tuning large diffusion models, we have not explored this direction in the current work. We consider this a promising avenue for future research.

\subsection{Analysis of Out-of-Distribution Data}
Sometimes, tracking videos may contain out-of-distribution content, such as infrared sequences, which fall outside the typical data distribution of models like SD-v1.5. To evaluate the robustness of our approach in such scenarios, we specifically tested GLAD on these sequences and analyzed its performance.
Specifically, we selected all infrared video sequences from the TNL2K test dataset and calculated the AUC. The results are shown in Table~\ref{tab:Infraed_auc}.
 
It can be observed that our method achieves a certain performance improvement, indicating that our method demonstrates good generalization ability and can still perform effective tracking on out-of-distribution data such as infrared imagery.

\begin{table*}[htbp]
\centering
\caption{Comparison of tracking performance with and without the template inpainting process.
The variant without inpainting excludes the diffusion-guided template enhancement but retains diffusion-based feature extraction.
}
\renewcommand{\arraystretch}{1.2} 
\setlength{\tabcolsep}{24pt}
\begin{tabular}{c | c | c | c}
    \toprule
    $AUC(\%)$ & LaSOT & TNL2K & OTB99-lang \\
    \midrule
    with template inpainting & \textbf{69.5} & \textbf{59.7} & \textbf{69.6} \\
    without template inpainting & 68.4 & 56.3 & 68.8 \\
    \bottomrule
\end{tabular}
\label{tab:TI}
\end{table*}

\begin{table}[htbp]
\centering
\caption{AUC of infraed sequences from TNL2K. All tests were conducted on a RTX 3090 GPU.}
\setlength{\tabcolsep}{15pt}
\renewcommand{\arraystretch}{1.2} 
\begin{tabular}{c | c}
    \toprule
    Method & Infraed Sequence AUC(\%) \\
    \midrule
    JointNLT & 51.7  \\
    VLT$_{\text{TT}}$ & 48.6 \\
    SeqTrack & 49.3 \\
    \midrule
    \textbf{GLAD-B256} & \textbf{53.2} \\
    \bottomrule
\end{tabular}
\label{tab:Infraed_auc}
\end{table}
 
\subsection{Analysis of Template Semantics}

From Table~\ref{tab:GLAD_SOTA_Comparison} we can find that our method outperforms most methods which leverage visual grounding to assist in vision-language tracking on several datasets.
To further analyze the effectiveness of our method, we will make a analysis from the restoration of semantic perspective of the template since our method is developed from the perspective of  the restoration of low template semantic.

In our setting, the term low semantic refers to visual inputs that suffer from degraded perceptual quality, resulting in reduced semantic clarity. Such degradation may arise from a wide range of real-world factors, including motion blur caused by rapid movement, insufficient illumination leading to low-contrast scenes, and low-resolution inputs where fine-grained features are underrepresented. Furthermore, environmental influences such as fog, haze, rain streaks, or smog can obscure important visual details, while sensor-level issues—such as high-ISO noise or compression artifacts—introduce distortions that further hinder interpretation. Partial occlusions or visual clutter can also mask key semantic elements. While these images may still contain meaningful content, the presence of such noise and distortion makes semantic extraction more challenging for models, thereby increasing the tracking difficulty for the model.

To facilitate analysis and quantitatively evaluate the improvement, we divide the semantics of the template into two components: the clarity of the tracking target and the relevance to the text description. The former refers to the resolution and detail sharpness of its visual information, while the latter refers to the semantic alignment between the visual features of the image and the linguistic features of the text. We will discuss the effectiveness of our method from these two perspectives separately.

\subsubsection{The Clarity of the Tracking Target}
\label{sec:vision}

The clarity of the tracking target is highly correlated with the overall performance of a tracking model, as a well-defined and visually detailed target provides more reliable cues for feature extraction and localization.
To investigate this relationship, we analyze the influence of template inpainting and further illustrate its impact through visual representations of the inpainting process.

\begin{table*}[htbp]
    \centering
    \caption{Quantitative statistics on mainstream datasets, including the number and proportion of High / Low-Semantic Frames and High / Low-Semantic Videos, as well as the Average Template Semantics values for each dataset.}
    \resizebox{\linewidth}{!}{
    \renewcommand{\arraystretch}{1.2} 
    \begin{tabular}{c|cc|cc|cc|cc|c}
    \toprule
       \multirow{3}{*}{\diagbox{Dataset}{Concepts}} & \multicolumn{2}{c|}{High-Semantic Frame} & \multicolumn{2}{c|}{Low-Semantic Frame} & \multicolumn{2}{c|}{High-Semantic Video} & \multicolumn{2}{c|}{Low-Semantic Video} & Average \\
       \cmidrule{2-9}
        & \multirow{2}{*}{Number} & \multirow{2}{*}{Ratio} & \multirow{2}{*}{Number} & \multirow{2}{*}{Ratio} & \multirow{2}{*}{Number} & \multirow{2}{*}{Ratio} & \multirow{2}{*}{Number} & \multirow{2}{*}{Ratio} & Template \\
        & & & & & & & & & Semantics \\
    \midrule
      LaSOT & 15764 & 46.2\% & 18352 & 53.8\% & 116 & 41.4\% & 164 & 58.6\% & 24.94 \\
      LaSOT$_{\text{ext}}$ & 8475 & 47.5\% & 9358 & 52.5\% & 71 & 47.3\% & 79 & 52.7\% & 24.56 \\
      TNL2K& 9326 & 36.7\% & 16097 & 63.3\% & 258 & 36.9\% & 442 & 63.1\% & 20.32 \\
      OTB99-lang & 661 & 46.7\% & 753 & 53.3\% & 20 & 41.7\% & 28 & 58.3\% & 21.98 \\
    \bottomrule
    \end{tabular}
    }
    \label{tab:Dataset_Analysis}
\end{table*}

Figure~\ref{fig: vis} presents several visualizations of the template inpainting results described in the Generative Diffusion Module (Section~\ref{GDM}).
The examples are drawn from LaSOT, OTB99-Lang, and TNL2K datasets.
It can be observed that after inpainting, the structural and textural details of the templates are effectively restored, which mitigates issues such as image blur, visual degradation, and ambiguous semantics, ultimately providing a clearer reference for tracking.

To quantitatively assess the contribution of template inpainting, we conduct an additional experiment by re-implementing a variant of our model that excludes the inpainting process but still employs diffusion-based feature extraction.
In this variant, only the U-Net component is used to extract template features without diffusion-guided fusion.

As shown in Table~\ref{tab:TI}, the GLAD variant without template inpainting experiences a noticeable performance drop in AUC on LaSOT and an even more significant degradation on TNL2K.
These results clearly demonstrate the effectiveness of the template inpainting strategy in enhancing template quality and semantic consistency, thereby validating that diffusion-based fusion of textual and visual information plays a crucial role in improving tracking robustness and precision.

\subsubsection{Relevance Analysis between Text and Video}
\label{sec:relevance}

Similarly, the relevance between texts and images is also a key factor affecting the performance of the model. To this end, we conducted an analysis on several vision-language tracking datasets. 
To quantitatively analyze the relevance between text and video on mainstream datasets, including LaSOT~\citep{fan2019lasot}, LaSOT$_{\text{ext}}$~\citep{fan2021lasot}, TNL2K~\citep{wang2021towards} and OTB99-lang~\citep{li2017tracking}.

In typical vision-language tracking settings, the template image and the corresponding textual description are expected to be semantically aligned and mutually complementary, ensuring that the linguistic cue accurately guides the tracker to identify and localize the intended target.  
Hence, the \textit{degree of semantic alignment} between the template and text can serve as an important indicator for evaluating the \textit{semantic validity, richness, and clarity} of the template representation.  
A well-aligned pair implies that the visual appearance of the template effectively reflects the semantic meaning expressed in the textual description, which facilitates better cross-modal understanding during tracking.

To quantify this semantic alignment, we employ the CLIP score, a widely recognized metric for measuring image–text correspondence.  
Specifically, the CLIP~\citep{radford2021learning} model jointly encodes visual and linguistic inputs into a shared embedding space, where semantic similarity is represented by the cosine similarity between the image and text embeddings.  
A higher CLIP score indicates that the template and textual description are more semantically consistent, suggesting that the template can better convey the intended meaning of the text and thus provide stronger guidance for the tracker.  
In our study, we use the CLIP score as a quantitative indicator of the semantic coherence between visual and textual modalities.

The formula for calculating the CLIP score is as follows:

\begin{equation}
\text{clip score}(I, T) = \frac{LS \cdot  \text{CLIP}(I) \cdot \text{CLIP}(T)}
{||\text{CLIP}(I)|| \cdot ||\text{CLIP}(T)||},
\end{equation}

where \(LS\) denotes the \textit{LogitScale}, a learned parameter in the original CLIP model that functions as a temperature coefficient.  
Here, \(I\) represents the image, \(T\) represents the text, and \(\text{CLIP}(I)\) and \(\text{CLIP}(T)\) are the corresponding feature embeddings extracted by the CLIP encoders.  
The operator ``\(\cdot\)’’ indicates the \textit{dot product} between vectors, while \(||\cdot||\) denotes their \textit{L2 norm}.  
It is also widely used by image caption models and generative models such as Imagen~\citep{saharia2022photorealistic} to evaluate the effectiveness of multi-modal models in matching textual descriptions with corresponding visual content.

\begin{table*}[htbp]
\centering
\caption{
CLIP score evaluation under different template degradations.
}
\renewcommand{\arraystretch}{1.2} 
\setlength{\tabcolsep}{16pt}
\begin{tabular}{c | c | c | c | c | c}
    \toprule
    Video & Raw & Blurring & Occlusion & Darkening & Color Jitter \\
    \midrule
    Clip Score & \textbf{24.94} & \textbf{21.19} & \textbf{19.86} & \textbf{24.15} & \textbf{23.54} \\
    \bottomrule
\end{tabular}
\label{tab:clip_eva}
\end{table*}

After quantifying the semantic correlation between images and textual descriptions, 
we can further extend this concept to the video-text level to evaluate the overall semantic consistency of videos to measure the global relevance between a video and its corresponding textual description, 
which reflects how well the textual semantics are preserved across the entire sequence. 
This assessment helps us better understand whether a tracker benefits from high-quality semantic templates when handling videos with different degrees of visual-textual alignment.

For each video, we first compute the CLIP score for the template image $f_{0}$ and then uniformly sample one frame from every 20 frames as the selected search image, denoted as $f_{20i}$. 
We compute the CLIP score for each sampled frame to obtain a set of semantic relevance scores:
\begin{equation}
    \{ S_{0}, S_{20}, S_{40}, \ldots, S_{20i}, \ldots \}.
\end{equation}
This score set represents the semantic correspondence between textual descriptions and video frames, 
enabling us to analyze the video-level semantic characteristics more effectively. 
To this end, we define several quantitative metrics as follows.

\noindent\textbf{High / Low-Semantic Frame:} 
For any selected search image $f_{20i}$ $(i > 0)$ in a video, if its score $S_{20i}$ is greater than the template score $S_{0}$, it is considered a high-semantic frame; 
otherwise, it is a low-semantic frame. 
This distinction helps evaluate the relative semantic adequacy of frames compared to the template.

\noindent\textbf{High / Low-Semantic Video:} 
We calculate the average value $\overline{S}$ of all frame-level scores $\{S_{20i}\}$. 
If $\overline{S}$ is higher than the template score $S_{0}$, the video is considered a high-semantic video; 
otherwise, it is regarded as a low-semantic video. 
This classification reflects the overall semantic relevance between the text and the visual content of the video.

\noindent\textbf{Average Template Semantics:} 
Finally, we compute the average of all template-level scores $S_{0}$ across the dataset, 
which serves as an indicator of the dataset’s overall semantic richness. 
This metric provides a global reference for comparing the semantic adequacy of templates and understanding their influence on tracking performance.

Based on the above concepts, we conduct a comprehensive statistical analysis on the test splits of LaSOT, TNL2K, OTB99-Lang, and LaSOT$_{\text{ext}}$, 
as summarized in Table~\ref{tab:Dataset_Analysis}. 
It can be observed that within most videos, the number of low-semantic frames is significantly greater than that of high-semantic ones, 
indicating that the semantic consistency between the text descriptions and the visual content is generally weak. 
Furthermore, the overall proportion of low-semantic videos exceeds that of high-semantic videos across all datasets, 
revealing that the textual descriptions in current benchmarks often fail to accurately align with or represent the visual targets throughout the entire sequence. 

Meanwhile, degradations such as blurring or occlusion further exacerbate the loss of semantic information in the template. To validate this, we conducted an additional experiment: we selected all template images from LaSOT based on our previous CLIP score evaluations and applied four types of degradations to each video—blurring, occlusion, darkening, and color jitter—to create controlled variants. We then recalculated the average CLIP score for each modified video using the same procedure. The results are shown in Table~\ref{tab:clip_eva}.The results are shown below. We observe a consistent decrease in the average CLIP score across all types of degradations, which not only validates our conclusion but also indicates that the CLIP score is a reasonably reliable measure of the overall semantic quality of a video.

\begin{table*}
    \centering
    \caption{Comparison of average template semantics before and after restoration with SD on different datasets.}
    \renewcommand{\arraystretch}{1.2} 
    \begin{tabular}{ c | c | c }
        \toprule
        Dataset & \makecell[c]{Original Template Semantics} & \makecell[c]{Template Semantics after SD} \\
        \midrule
        LaSOT & 24.94 & 25.78 \\
        LaSOT-ext & 24.56 & 26.87 \\
        TNL2K & 20.32 & 23.09 \\
        OTB99-lang & 21.98 & 22.48 \\
        \bottomrule
    \end{tabular}
    \label{tab:Clip_Score}
\end{table*}

This observation implies that directly introducing textual features into the search-region representation may lead to semantic confusion and unstable attention, 
as the search frames themselves often lack sufficient semantic correspondence with the text. 
In contrast, the template and the text both serve as descriptions of the same target object — one in visual form and the other in linguistic form. 
Therefore, enhancing the interaction between template features and textual features allows the model to more effectively consolidate multimodal cues about the target’s appearance and attributes, 
producing a semantically richer and more consistent target representation. 
This design philosophy underpins the GLAD framework, which prioritizes generative refinement of the template through text-guided diffusion before performing multimodal fusion. 
By doing so, GLAD establishes a more reliable target prior, thereby improving the accuracy and robustness of subsequent tracking.

In addition, to further quantify the semantic enhancement provided by our approach, 
we compute the CLIP score for template images before and after template inpainting using Stable Diffusion (SD) 
and compare the average template semantics across datasets. 
The results, shown in Table~\ref{tab:Clip_Score}, 
demonstrate a consistent improvement in the post-restoration CLIP scores. 
This indicates that our diffusion-based inpainting effectively enhances the semantic expressiveness of templates, 
bringing them closer to the text descriptions and improving the overall multimodal alignment. 
Consequently, these refined templates contribute to more reliable visual-textual associations, 
which are crucial for achieving stable and accurate tracking performance in vision-language scenarios.

\section{Conclusion}\label{sec: conclusion}

In this paper, we have proposed a new vision-language tracking method GLAD. It's based on the novel generative multi-modal fusion paradigm and comprises the Generative Diffusion Module and the cascaded Multi-Modal Decoder. Textual information is firstly fused with template features for semantic restoration. Search features are then interacted with the generated multi-modal features for semantic enhancement.
The experiments have demonstrated that our method effectively restores low-semantic templates and enhances tracking performance.
Besides, we also present a comprehensive analysis of template semantics to further confirm the effectiveness and theoretical reliability of our method.
As a result, our GLAD achieves excellent performance in both accuracy and inference speed. We hope GLAD can provide a new option for the tracking community in the future.

\backmatter

% \bmhead{Supplementary information}

% If your article has accompanying supplementary file/s please state so here. 

% Authors reporting data from electrophoretic gels and blots should supply the full unprocessed scans for key as part of their Supplementary information. This may be requested by the editorial team/s if it is missing.

% Please refer to Journal-level guidance for any specific requirements.

% \bmhead{Acknowledgements}

% Acknowledgements are not compulsory. Where included they should be brief. Grant or contribution numbers may be acknowledged.

% Please refer to Journal-level guidance for any specific requirements.

\paragraph{\bf Acknowledgement.} 
This work is supported by the National Natural Science Foundation of China (Grant No. 62402211) and the Natural Science Foundation of Jiangsu Province (Grant No. BK20241248).

\paragraph{\bf Conflict of interest statement.} 
There is no specific conflict of interest statement with this manuscript

\paragraph{\bf Compliance of ethical standard statement.} 
There is no specific compliance of ethical standard statement with this manuscript

\paragraph{\bf Informed consent.} 
There is no specific informed consent with this manuscript

\paragraph{\bf Funding Information.} 
This work is supported by the National Natural Science Foundation of China (Grant No. 62402211) and the Natural Science Foundation of Jiangsu Province (Grant No. BK20241248).

\paragraph{\bf Data Availability.} 
This manuscript develops its me-thod based on the publicly available datasets: GOT-10k~\citep{huang2019got}, LaSOT~\citep{fan2019lasot}, OTB99-lang~\citep{li2017tracking}, TNL2K~\citep{wang2021towards}, TrackingNet\citep{muller2018trackingnet}, RefCOCOg~\citep{DBLP:conf/cvpr/MaoHTCY016}. There is no specific associated data with this manuscript and the code included in this study are available upon request by contact with the corresponding author.

\bibliography{sn-bibliography}% common bib file

@String(CVPR= {IEEE Conf. Comput. Vis. Pattern Recog.})

@String(ICCV= {Int. Conf. Comput. Vis.})

@String(ECCV= {Eur. Conf. Comput. Vis.})

@String(IJCAI = {IJCAI})

@String(AAAI = {AAAI})

@String(CVPR  = {CVPR})

@String(ICCV  = {ICCV})

@String(ECCV  = {ECCV})

@inproceedings{feng2021siamese,
  title={Siamese natural language tracker: Tracking by natural language descriptions with siamese trackers},
  author={Feng, Qi and Ablavsky, Vitaly and Bai, Qinxun and Sclaroff, Stan},
  booktitle={Proceedings of the IEEE/CVF conference on computer vision and pattern recognition},
  pages={5851--5860},
  year={2021}
}

@article{devlin2018bert,
  title={Bert: Pre-training of deep bidirectional transformers for language understanding},
  author={Devlin, Jacob and Chang, Ming-Wei and Lee, Kenton and Toutanova, Kristina},
  journal={arXiv preprint arXiv:1810.04805},
  year={2018}
}

@article{brown2020language,
  title={Language models are few-shot learners},
  author={Brown, Tom and Mann, Benjamin and Ryder, Nick and Subbiah, Melanie and Kaplan, Jared D and Dhariwal, Prafulla and Neelakantan, Arvind and Shyam, Pranav and Sastry, Girish and Askell, Amanda and others},
  journal={Advances in neural information processing systems},
  volume={33},
  pages={1877--1901},
  year={2020}
}

@inproceedings{rombach2022high,
  title={High-resolution image synthesis with latent diffusion models},
  author={Rombach, Robin and Blattmann, Andreas and Lorenz, Dominik and Esser, Patrick and Ommer, Bj{\"o}rn},
  booktitle={Proceedings of the IEEE/CVF conference on computer vision and pattern recognition},
  pages={10684--10695},
  year={2022}
}

@inproceedings{chen2023diffusiondet,
  title={Diffusiondet: Diffusion model for object detection},
  author={Chen, Shoufa and Sun, Peize and Song, Yibing and Luo, Ping},
  booktitle={Proceedings of the IEEE/CVF International Conference on Computer Vision},
  pages={19830--19843},
  year={2023}
}

@inproceedings{pnvr2023ld,
  title={Ld-znet: A latent diffusion approach for text-based image segmentation},
  author={Pnvr, Koutilya and Singh, Bharat and Ghosh, Pallabi and Siddiquie, Behjat and Jacobs, David},
  booktitle={Proceedings of the IEEE/CVF International Conference on Computer Vision},
  pages={4157--4168},
  year={2023}
}

@inproceedings{wang2021towards,
  title={Towards more flexible and accurate object tracking with natural language: Algorithms and benchmark},
  author={Wang, Xiao and Shu, Xiujun and Zhang, Zhipeng and Jiang, Bo and Wang, Yaowei and Tian, Yonghong and Wu, Feng},
  booktitle={Proceedings of the IEEE/CVF conference on computer vision and pattern recognition},
  pages={13763--13773},
  year={2021}
}

@inproceedings{li2017tracking,
  title={Tracking by natural language specification},
  author={Li, Zhenyang and Tao, Ran and Gavves, Efstratios and Snoek, Cees GM and Smeulders, Arnold WM},
  booktitle={Proceedings of the IEEE conference on computer vision and pattern recognition},
  pages={6495--6503},
  year={2017}
}

@article{guo2022divert,
  title={Divert more attention to vision-language tracking},
  author={Guo, Mingzhe and Zhang, Zhipeng and Fan, Heng and Jing, Liping},
  journal={Advances in Neural Information Processing Systems},
  volume={35},
  pages={4446--4460},
  year={2022}
}

@inproceedings{ye2022joint,
  title={Joint feature learning and relation modeling for tracking: A one-stream framework},
  author={Ye, Botao and Chang, Hong and Ma, Bingpeng and Shan, Shiguang and Chen, Xilin},
  booktitle={European conference on computer vision},
  pages={341--357},
  year={2022},
  organization={Springer}
}

@inproceedings{zhou2023joint,
  title={Joint visual grounding and tracking with natural language specification},
  author={Zhou, Li and Zhou, Zikun and Mao, Kaige and He, Zhenyu},
  booktitle={Proceedings of the IEEE/CVF conference on computer vision and pattern recognition},
  pages={23151--23160},
  year={2023}
}

@article{zhao2023transformer,
  title={Transformer vision-language tracking via proxy token guided cross-modal fusion},
  author={Zhao, Haojie and Wang, Xiao and Wang, Dong and Lu, Huchuan and Ruan, Xiang},
  journal={Pattern Recognition Letters},
  volume={168},
  pages={10--16},
  year={2023},
  publisher={Elsevier}
}

@inproceedings{fan2019lasot,
  title={Lasot: A high-quality benchmark for large-scale single object tracking},
  author={Fan, Heng and Lin, Liting and Yang, Fan and Chu, Peng and Deng, Ge and Yu, Sijia and Bai, Hexin and Xu, Yong and Liao, Chunyuan and Ling, Haibin},
  booktitle={Proceedings of the IEEE/CVF conference on computer vision and pattern recognition},
  pages={5374--5383},
  year={2019}
}

@article{fan2021lasot,
  title={Lasot: A high-quality large-scale single object tracking benchmark},
  author={Fan, Heng and Bai, Hexin and Lin, Liting and Yang, Fan and Chu, Peng and Deng, Ge and Yu, Sijia and Harshit and Huang, Mingzhen and Liu, Juehuan and others},
  journal={International Journal of Computer Vision},
  volume={129},
  pages={439--461},
  year={2021},
  publisher={Springer}
}

@article{huang2019got,
  title={Got-10k: A large high-diversity benchmark for generic object tracking in the wild},
  author={Huang, Lianghua and Zhao, Xin and Huang, Kaiqi},
  journal={IEEE transactions on pattern analysis and machine intelligence},
  volume={43},
  number={5},
  pages={1562--1577},
  year={2019},
  publisher={IEEE}
}

@article{song2023consistency,
  title={Consistency models},
  author={Song, Yang and Dhariwal, Prafulla and Chen, Mark and Sutskever, Ilya},
  journal={arXiv preprint arXiv:2303.01469},
  year={2023}
}

@article{luo2023lcm,
  title={Lcm-lora: A universal stable-diffusion acceleration module},
  author={Luo, Simian and Tan, Yiqin and Patil, Suraj and Gu, Daniel and von Platen, Patrick and Passos, Apolin{\'a}rio and Huang, Longbo and Li, Jian and Zhao, Hang},
  journal={arXiv preprint arXiv:2311.05556},
  year={2023}
}

@inproceedings{cui2022mixformer,
  title={Mixformer: End-to-end tracking with iterative mixed attention},
  author={Cui, Yutao and Jiang, Cheng and Wang, Limin and Wu, Gangshan},
  booktitle={Proceedings of the IEEE/CVF conference on computer vision and pattern recognition},
  pages={13608--13618},
  year={2022}
}

@article{vaswani2017attention,
  title={Attention is all you need},
  author={Vaswani, Ashish and Shazeer, Noam and Parmar, Niki and Uszkoreit, Jakob and Jones, Llion and Gomez, Aidan N and Kaiser, {\L}ukasz and Polosukhin, Illia},
  journal={Advances in neural information processing systems},
  volume={30},
  year={2017}
}

@inproceedings{radford2021learning,
  title={Learning transferable visual models from natural language supervision},
  author={Radford, Alec and Kim, Jong Wook and Hallacy, Chris and Ramesh, Aditya and Goh, Gabriel and Agarwal, Sandhini and Sastry, Girish and Askell, Amanda and Mishkin, Pamela and Clark, Jack and others},
  booktitle={International conference on machine learning},
  pages={8748--8763},
  year={2021},
  organization={PMLR}
}

@article{ho2020denoising,
  title={Denoising diffusion probabilistic models},
  author={Ho, Jonathan and Jain, Ajay and Abbeel, Pieter},
  journal={Advances in neural information processing systems},
  volume={33},
  pages={6840--6851},
  year={2020}
}

@inproceedings{sohl2015deep,
  title={Deep unsupervised learning using nonequilibrium thermodynamics},
  author={Sohl-Dickstein, Jascha and Weiss, Eric and Maheswaranathan, Niru and Ganguli, Surya},
  booktitle={International conference on machine learning},
  pages={2256--2265},
  year={2015},
  organization={PMLR}
}

@article{song2019generative,
  title={Generative modeling by estimating gradients of the data distribution},
  author={Song, Yang and Ermon, Stefano},
  journal={Advances in neural information processing systems},
  volume={32},
  year={2019}
}

@article{song2020score,
  title={Score-based generative modeling through stochastic differential equations},
  author={Song, Yang and Sohl-Dickstein, Jascha and Kingma, Diederik P and Kumar, Abhishek and Ermon, Stefano and Poole, Ben},
  journal={arXiv preprint arXiv:2011.13456},
  year={2020}
}

@article{song2020improved,
  title={Improved techniques for training score-based generative models},
  author={Song, Yang and Ermon, Stefano},
  journal={Advances in neural information processing systems},
  volume={33},
  pages={12438--12448},
  year={2020}
}

@article{song2021maximum,
  title={Maximum likelihood training of score-based diffusion models},
  author={Song, Yang and Durkan, Conor and Murray, Iain and Ermon, Stefano},
  journal={Advances in neural information processing systems},
  volume={34},
  pages={1415--1428},
  year={2021}
}

@article{karras2022elucidating,
  title={Elucidating the design space of diffusion-based generative models},
  author={Karras, Tero and Aittala, Miika and Aila, Timo and Laine, Samuli},
  journal={Advances in Neural Information Processing Systems},
  volume={35},
  pages={26565--26577},
  year={2022}
}

@inproceedings{muller2018trackingnet,
  title={Trackingnet: A large-scale dataset and benchmark for object tracking in the wild},
  author={Muller, Matthias and Bibi, Adel and Giancola, Silvio and Alsubaihi, Salman and Ghanem, Bernard},
  booktitle={Proceedings of the European conference on computer vision (ECCV)},
  pages={300--317},
  year={2018}
}

@inproceedings{glorot2010understanding,
  title={Understanding the difficulty of training deep feedforward neural networks},
  author={Glorot, Xavier and Bengio, Yoshua},
  booktitle={Proceedings of the thirteenth international conference on artificial intelligence and statistics},
  pages={249--256},
  year={2010},
  organization={JMLR Workshop and Conference Proceedings}
}

@article{loshchilov2017decoupled,
  title={Decoupled weight decay regularization},
  author={Loshchilov, Ilya and Hutter, Frank},
  journal={arXiv preprint arXiv:1711.05101},
  year={2017}
}

@inproceedings{rasley2020deepspeed,
  title={Deepspeed: System optimizations enable training deep learning models with over 100 billion parameters},
  author={Rasley, Jeff and Rajbhandari, Samyam and Ruwase, Olatunji and He, Yuxiong},
  booktitle={Proceedings of the 26th ACM SIGKDD International Conference on Knowledge Discovery \& Data Mining},
  pages={3505--3506},
  year={2020}
}

@inproceedings{rajbhandari2020zero,
  title={Zero: Memory optimizations toward training trillion parameter models},
  author={Rajbhandari, Samyam and Rasley, Jeff and Ruwase, Olatunji and He, Yuxiong},
  booktitle={SC20: International Conference for High Performance Computing, Networking, Storage and Analysis},
  pages={1--16},
  year={2020},
  organization={IEEE}
}

@inproceedings{chen2021transformer,
  title={Transformer tracking},
  author={Chen, Xin and Yan, Bin and Zhu, Jiawen and Wang, Dong and Yang, Xiaoyun and Lu, Huchuan},
  booktitle={Proceedings of the IEEE/CVF conference on computer vision and pattern recognition},
  pages={8126--8135},
  year={2021}
}

@inproceedings{zhang2020ocean,
  title={Ocean: Object-aware anchor-free tracking},
  author={Zhang, Zhipeng and Peng, Houwen and Fu, Jianlong and Li, Bing and Hu, Weiming},
  booktitle={Computer Vision--ECCV 2020: 16th European Conference, Glasgow, UK, August 23--28, 2020, Proceedings, Part XXI 16},
  pages={771--787},
  year={2020},
  organization={Springer}
}

@inproceedings{bhat2019learning,
  title={Learning discriminative model prediction for tracking},
  author={Bhat, Goutam and Danelljan, Martin and Gool, Luc Van and Timofte, Radu},
  booktitle={Proceedings of the IEEE/CVF international conference on computer vision},
  pages={6182--6191},
  year={2019}
}

@inproceedings{wei2023autoregressive,
  title={Autoregressive visual tracking},
  author={Wei, Xing and Bai, Yifan and Zheng, Yongchao and Shi, Dahu and Gong, Yihong},
  booktitle={Proceedings of the IEEE/CVF Conference on Computer Vision and Pattern Recognition},
  pages={9697--9706},
  year={2023}
}

@inproceedings{feng2020real,
  title={Real-time visual object tracking with natural language description},
  author={Feng, Qi and Ablavsky, Vitaly and Bai, Qinxun and Li, Guorong and Sclaroff, Stan},
  booktitle={Proceedings of the IEEE/CVF Winter Conference on Applications of Computer Vision},
  pages={700--709},
  year={2020}
}

@inproceedings{yang2019fast,
  title={A fast and accurate one-stage approach to visual grounding},
  author={Yang, Zhengyuan and Gong, Boqing and Wang, Liwei and Huang, Wenbing and Yu, Dong and Luo, Jiebo},
  booktitle={Proceedings of the IEEE/CVF international conference on computer vision},
  pages={4683--4693},
  year={2019}
}

@inproceedings{huang2021look,
  title={Look before you leap: Learning landmark features for one-stage visual grounding},
  author={Huang, Binbin and Lian, Dongze and Luo, Weixin and Gao, Shenghua},
  booktitle={Proceedings of the IEEE/CVF conference on computer vision and pattern recognition},
  pages={16888--16897},
  year={2021}
}

@article{yang2020grounding,
  title={Grounding-tracking-integration},
  author={Yang, Zhengyuan and Kumar, Tushar and Chen, Tianlang and Su, Jingsong and Luo, Jiebo},
  journal={IEEE Transactions on Circuits and Systems for Video Technology},
  volume={31},
  number={9},
  pages={3433--3443},
  year={2020},
  publisher={IEEE}
}

@inproceedings{li2022cross,
  title={Cross-modal target retrieval for tracking by natural language},
  author={Li, Yihao and Yu, Jun and Cai, Zhongpeng and Pan, Yuwen},
  booktitle={Proceedings of the IEEE/CVF Conference on Computer Vision and Pattern Recognition},
  pages={4931--4940},
  year={2022}
}

@article{CenterNet,
  author    = {Xingyi Zhou and
               Dequan Wang and
               Philipp Kr{\"{a}}henb{\"{u}}hl},
  title     = {Objects as Points},
  journal   = {CoRR},
  volume    = {abs/1904.07850},
  year      = {2019}
}

@inproceedings{he2016deep,
  title={Deep residual learning for image recognition},
  author={He, Kaiming and Zhang, Xiangyu and Ren, Shaoqing and Sun, Jian},
  booktitle={Proceedings of the IEEE conference on computer vision and pattern recognition},
  pages={770--778},
  year={2016}
}

@article{dosovitskiy2020image,
  title={An image is worth 16x16 words: Transformers for image recognition at scale},
  author={Dosovitskiy, Alexey and Beyer, Lucas and Kolesnikov, Alexander and Weissenborn, Dirk and Zhai, Xiaohua and Unterthiner, Thomas and Dehghani, Mostafa and Minderer, Matthias and Heigold, Georg and Gelly, Sylvain and others},
  journal={arXiv preprint arXiv:2010.11929},
  year={2020}
}

@inproceedings{shao2024context,
  title={Context-Aware Integration of Language and Visual References for Natural Language Tracking},
  author={Shao, Yanyan and He, Shuting and Ye, Qi and Feng, Yuchao and Luo, Wenhan and Chen, Jiming},
  booktitle={Proceedings of the IEEE/CVF Conference on Computer Vision and Pattern Recognition},
  pages={19208--19217},
  year={2024}
}

@article{kingma2013auto,
  title={Auto-encoding variational bayes},
  author={Kingma, Diederik P and Welling, Max},
  journal={arXiv preprint arXiv:1312.6114},
  year={2013}
}

@inproceedings{rezende2014stochastic,
  title={Stochastic backpropagation and approximate inference in deep generative models},
  author={Rezende, Danilo Jimenez and Mohamed, Shakir and Wierstra, Daan},
  booktitle={International conference on machine learning},
  pages={1278--1286},
  year={2014},
  organization={PMLR}
}

@article{dhariwal2021diffusion,
  title={Diffusion models beat gans on image synthesis},
  author={Dhariwal, Prafulla and Nichol, Alexander},
  journal={Advances in neural information processing systems},
  volume={34},
  pages={8780--8794},
  year={2021}
}

@inproceedings{ronneberger2015u,
  title={U-net: Convolutional networks for biomedical image segmentation},
  author={Ronneberger, Olaf and Fischer, Philipp and Brox, Thomas},
  booktitle={Medical image computing and computer-assisted intervention--MICCAI 2015: 18th international conference, Munich, Germany, October 5-9, 2015, proceedings, part III 18},
  pages={234--241},
  year={2015},
  organization={Springer}
}

@article{saharia2022photorealistic,
  title={Photorealistic text-to-image diffusion models with deep language understanding},
  author={Saharia, Chitwan and Chan, William and Saxena, Saurabh and Li, Lala and Whang, Jay and Denton, Emily L and Ghasemipour, Kamyar and Gontijo Lopes, Raphael and Karagol Ayan, Burcu and Salimans, Tim and others},
  journal={Advances in neural information processing systems},
  volume={35},
  pages={36479--36494},
  year={2022}
}

@inproceedings{Staple,
  author    = {Luca Bertinetto and
               Jack Valmadre and
               Stuart Golodetz and
               Ondrej Miksik and
               Philip H. S. Torr},
  title     = {Staple: Complementary Learners for Real-Time Tracking},
  booktitle = {{CVPR}},
  pages     = {1401--1409},
  publisher = {{IEEE} Computer Society},
  year      = {2016}
}

@inproceedings{BhatJDKF18,
  author    = {Goutam Bhat and
               Joakim Johnander and
               Martin Danelljan and
               Fahad Shahbaz Khan and
               Michael Felsberg},
  title     = {Unveiling the Power of Deep Tracking},
  booktitle = {{ECCV} {(2)}},
  series    = {Lecture Notes in Computer Science},
  volume    = {11206},
  pages     = {493--509},
  publisher = {Springer},
  year      = {2018}
}

@inproceedings{ATOM,
  author    = {Martin Danelljan and
               Goutam Bhat and
               Fahad Shahbaz Khan and
               Michael Felsberg},
  title     = {{ATOM:} Accurate Tracking by Overlap Maximization},
  booktitle = {{CVPR}},
  pages     = {4660--4669},
  publisher = {Computer Vision Foundation / {IEEE}},
  year      = {2019}
}

@inproceedings{Song0WGBZSL018,
  author    = {Yibing Song and
               Chao Ma and
               Xiaohe Wu and
               Lijun Gong and
               Linchao Bao and
               Wangmeng Zuo and
               Chunhua Shen and
               Rynson W. H. Lau and
               Ming{-}Hsuan Yang},
  title     = {{VITAL:} VIsual Tracking via Adversarial Learning},
  booktitle = {{CVPR}},
  pages     = {8990--8999},
  publisher = {Computer Vision Foundation / {IEEE} Computer Society},
  year      = {2018}
}

@inproceedings{WangZLWTB21,
  author    = {Zhongdao Wang and
               Hengshuang Zhao and
               Ya{-}Li Li and
               Shengjin Wang and
               Philip H. S. Torr and
               Luca Bertinetto},
  title     = {Do Different Tracking Tasks Require Different Appearance Models?},
  booktitle = {NeurIPS},
  pages     = {726--738},
  year      = {2021}
}

@inproceedings{DBLP:conf/cvpr/ChenPWLH23,
  author       = {Xin Chen and
                  Houwen Peng and
                  Dong Wang and
                  Huchuan Lu and
                  Han Hu},
  title        = {SeqTrack: Sequence to Sequence Learning for Visual Object Tracking},
  booktitle    = {{CVPR}},
  pages        = {14572--14581},
  publisher    = {{IEEE}},
  year         = {2023}
}

@inproceedings{DBLP:conf/cvpr/LugmayrDRYTG22,
  author       = {Andreas Lugmayr and
                  Martin Danelljan and
                  Andr{\'{e}}s Romero and
                  Fisher Yu and
                  Radu Timofte and
                  Luc Van Gool},
  title        = {RePaint: Inpainting using Denoising Diffusion Probabilistic Models},
  booktitle    = {{CVPR}},
  pages        = {11451--11461},
  publisher    = {{IEEE}},
  year         = {2022}
}

@inproceedings{STARK,
  author    = {Bin Yan and
               Houwen Peng and
               Jianlong Fu and
               Dong Wang and
               Huchuan Lu},
  title     = {Learning Spatio-Temporal Transformer for Visual Tracking},
  booktitle = {{ICCV}},
  pages     = {10428--10437},
  publisher = {{IEEE}},
  year      = {2021}
}

@inproceedings{AiATrack,
  author    = {Shenyuan Gao and
               Chunluan Zhou and
               Chao Ma and
               Xinggang Wang and
               Junsong Yuan},
  title     = {AiATrack: Attention in Attention for Transformer Visual Tracking},
  booktitle = {{ECCV} {(22)}},
  series    = {Lecture Notes in Computer Science},
  volume    = {13682},
  pages     = {146--164},
  publisher = {Springer},
  year      = {2022}
}

@inproceedings{SimTrack,
  author    = {Boyu Chen and
               Peixia Li and
               Lei Bai and
               Lei Qiao and
               Qiuhong Shen and
               Bo Li and
               Weihao Gan and
               Wei Wu and
               Wanli Ouyang},
  title     = {Backbone is All Your Need: {A} Simplified Architecture for Visual
               Object Tracking},
  booktitle = {{ECCV} {(22)}},
  series    = {Lecture Notes in Computer Science},
  volume    = {13682},
  pages     = {375--392},
  publisher = {Springer},
  year      = {2022}
}

@inproceedings{DBLP:conf/ijcai/ZhangZLMS24,
  author       = {Guangtong Zhang and
                  Bineng Zhong and
                  Qihua Liang and
                  Zhiyi Mo and
                  Shuxiang Song},
  title        = {Diffusion Mask-Driven Visual-language Tracking},
  booktitle    = {{IJCAI}},
  pages        = {1652--1660},
  publisher    = {ijcai.org},
  year         = {2024}
}

@inproceedings{zhang2023all,
  title={All in one: Exploring unified vision-language tracking with multi-modal alignment},
  author={Zhang, Chunhui and Sun, Xin and Yang, Yiqian and Liu, Li and Liu, Qiong and Zhou, Xi and Wang, Yanfeng},
  booktitle={Proceedings of the 31st ACM International Conference on Multimedia},
  pages={5552--5561},
  year={2023}
}

@inproceedings{DBLP:conf/cvpr/HeCXLDG22,
  author       = {Kaiming He and
                  Xinlei Chen and
                  Saining Xie and
                  Yanghao Li and
                  Piotr Doll{\'{a}}r and
                  Ross B. Girshick},
  title        = {Masked Autoencoders Are Scalable Vision Learners},
  booktitle    = {{CVPR}},
  pages        = {15979--15988},
  publisher    = {{IEEE}},
  year         = {2022}
}

@article{zheng2023toward,
  title={Toward unified token learning for vision-language tracking},
  author={Zheng, Yaozong and Zhong, Bineng and Liang, Qihua and Li, Guorong and Ji, Rongrong and Li, Xianxian},
  journal={IEEE Transactions on Circuits and Systems for Video Technology},
  volume={34},
  number={4},
  pages={2125--2135},
  year={2023},
  publisher={IEEE}
}

@inproceedings{ma2023tracking,
  title={Tracking by natural language specification with long short-term context decoupling},
  author={Ma, Ding and Wu, Xiangqian},
  booktitle={Proceedings of the IEEE/CVF International Conference on Computer Vision},
  pages={14012--14021},
  year={2023}
}

@inproceedings{DBLP:conf/nips/SunYCZHLLW24,
  author       = {Yiming Sun and
                  Fan Yu and
                  Shaoxiang Chen and
                  Yu Zhang and
                  Junwei Huang and
                  Yang Li and
                  Chenhui Li and
                  Changbo Wang},
  title        = {ChatTracker: Enhancing Visual Tracking Performance via Chatting with
                  Multimodal Large Language Model},
  booktitle    = {NeurIPS},
  year         = {2024}
}

@inproceedings{DBLP:conf/cvpr/MaoHTCY016,
  author       = {Junhua Mao and
                  Jonathan Huang and
                  Alexander Toshev and
                  Oana Camburu and
                  Alan L. Yuille and
                  Kevin Murphy},
  title        = {Generation and Comprehension of Unambiguous Object Descriptions},
  booktitle    = {{CVPR}},
  pages        = {11--20},
  publisher    = {{IEEE} Computer Society},
  year         = {2016}
}

@InProceedings{Cai_2023_ICCV,
    author    = {Cai, Yidong and Liu, Jie and Tang, Jie and Wu, Gangshan},
    title     = {Robust Object Modeling for Visual Tracking},
    booktitle = {Proceedings of the IEEE/CVF International Conference on Computer Vision (ICCV)},
    month     = {October},
    year      = {2023},
    pages     = {9589-9600}
}

@inproceedings{chen2025sutrack,
  title={SUTrack: Towards simple and unified single object tracking},
  author={Chen, Xin and Kang, Ben and Geng, Wanting and Zhu, Jiawen and Liu, Yi and Wang, Dong and Lu, Huchuan},
  booktitle={Proceedings of the AAAI Conference on Artificial Intelligence},
  volume={39},
  number={2},
  pages={2239--2247},
  year={2025}
}

@article{ge2025improving,
  title={Improving Object Detection Models via LLM-Based Training Data Synthesis},
  author={Ge, Shi-Ran and Cao, Jie and He, Ran},
  journal={International Journal of Computer Vision},
  pages={1--16},
  year={2025},
  publisher={Springer}
}

@article{li2025diffusion,
  title={Diffusion models for image restoration and enhancement: a comprehensive survey},
  author={Li, Xin and Ren, Yulin and Jin, Xin and Lan, Cuiling and Wang, Xingrui and Zeng, Wenjun and Wang, Xinchao and Chen, Zhibo},
  journal={International Journal of Computer Vision},
  pages={1--31},
  year={2025},
  publisher={Springer}
}

@article{rassin2022dalle,
  title={DALLE-2 is seeing double: Flaws in word-to-concept mapping in Text2Image models},
  author={Rassin, Royi and Ravfogel, Shauli and Goldberg, Yoav},
  journal={arXiv preprint arXiv:2210.10606},
  year={2022}
}

@article{russakovsky2015imagenet,
  title={Imagenet large scale visual recognition challenge},
  author={Russakovsky, Olga and Deng, Jia and Su, Hao and Krause, Jonathan and Satheesh, Sanjeev and Ma, Sean and Huang, Zhiheng and Karpathy, Andrej and Khosla, Aditya and Bernstein, Michael and others},
  journal={International journal of computer vision},
  volume={115},
  number={3},
  pages={211--252},
  year={2015},
  publisher={Springer}
}

@article{cheng2025breaking,
  title={Breaking the limits of reliable prediction via generated data},
  author={Cheng, Zhen and Zhu, Fei and Zhang, Xu-Yao and Liu, Cheng-Lin},
  journal={International Journal of Computer Vision},
  volume={133},
  number={3},
  pages={1195--1221},
  year={2025},
  publisher={Springer}
}

@article{zhao2020layout2image,
  title={Layout2image: Image generation from layout},
  author={Zhao, Bo and Yin, Weidong and Meng, Lili and Sigal, Leonid},
  journal={International journal of computer vision},
  volume={128},
  number={10},
  pages={2418--2435},
  year={2020},
  publisher={Springer}
}

@article{quan2024deep,
  title={Deep learning-based image and video inpainting: A survey},
  author={Quan, Weize and Chen, Jiaxi and Liu, Yanli and Yan, Dong-Ming and Wonka, Peter},
  journal={International Journal of Computer Vision},
  volume={132},
  number={7},
  pages={2367--2400},
  year={2024},
  publisher={Springer}
}

@article{cui2023mixformerv2,
  title={Mixformerv2: Efficient fully transformer tracking},
  author={Cui, Yutao and Song, Tianhui and Wu, Gangshan and Wang, Limin},
  journal={Advances in neural information processing systems},
  volume={36},
  pages={58736--58751},
  year={2023}
}

\end{document}